\documentclass[runningheads]{llncs}

\usepackage{eccv}

\usepackage{eccvabbrv}

\usepackage{graphicx}
\usepackage{booktabs}
\usepackage{multirow}

\usepackage{subcaption}
\usepackage[accsupp]{axessibility}  %

\usepackage[pagebackref,breaklinks,colorlinks,citecolor=eccvblue]{hyperref}

\usepackage{orcidlink}
\usepackage{array}
\newcolumntype{L}[1]{>{\raggedright\let\newline\\\arraybackslash\hspace{0pt}}m{#1}}
\newcolumntype{C}[1]{>{\centering\let\newline\\\arraybackslash\hspace{0pt}}m{#1}}
\newcolumntype{R}[1]{>{\raggedleft\let\newline\\\arraybackslash\hspace{0pt}}m{#1}}

\begin{document}

\title{RAVE: Residual Vector Embedding for CLIP-Guided Backlit Image Enhancement}

\titlerunning{RAVE: Residual Vector Embedding}

\author{Tatiana Gaintseva\inst{1, 2}\orcidlink{0000-0003-4492-9789} \and
Martin Benning\inst{3}\orcidlink{0000-0002-6203-1350} \and
Gregory Slabaugh\inst{1, 2}\orcidlink{0000-0003-4060-5226}}

\authorrunning{T.~Gaintseva et al.}

\institute{Digital Environment Research Institute,
Queen Mary University of London, UK \and
School of Electronic Engineering and Computer Science, Queen Mary University of London, UK
\email{\{t.gaintseva, g.slabaugh\}@qmul.ac.uk}\\
\and
Department of Computer Science,
University College London, UK\\
\email{martin.benning@ucl.ac.uk}}

\maketitle

\vspace{-20em}
\underline{Accepted to ECCV 2024}
\vspace{20em}

\begin{abstract}
  In this paper we propose a novel modification of Contrastive Language-Image Pre-Training (CLIP) guidance for the task of backlit image enhancement. Our work builds on the state-of-the-art CLIP-LIT approach, which learns a prompt pair by constraining the text-image similarity between a prompt (negative/positive sample) and a corresponding image (backlit image/well-lit image) in the CLIP embedding space. Learned prompts then guide an image enhancement network.
Based on the CLIP-LIT framework, we propose two novel methods for CLIP guidance. First, we show that instead of tuning prompts in the space of text embeddings, it is possible to directly tune their embeddings in the latent space without any loss in quality. This accelerates training and potentially enables the use of additional encoders that do not have a text encoder. Second, 
we propose a novel approach that does not require any prompt tuning. Instead, based on CLIP embeddings of backlit and well-lit images from training data, we compute the residual vector in the embedding space as a simple difference between the mean embeddings of the well-lit and backlit images. This vector then guides the enhancement network during training, pushing a backlit image towards the space of well-lit images. This approach further dramatically reduces training time, stabilizes training and produces high quality enhanced images without artifacts. Additionally, we show that residual vectors can be interpreted, revealing biases in training data, and thereby enabling potential bias correction. Code is available at \url{https://github.com/Atmyre/RAVE}
\end{abstract}

\section{Introduction}
\label{sec:intro}

\begin{figure}[tb]
  \centering
      \includegraphics[height=6.3cm]{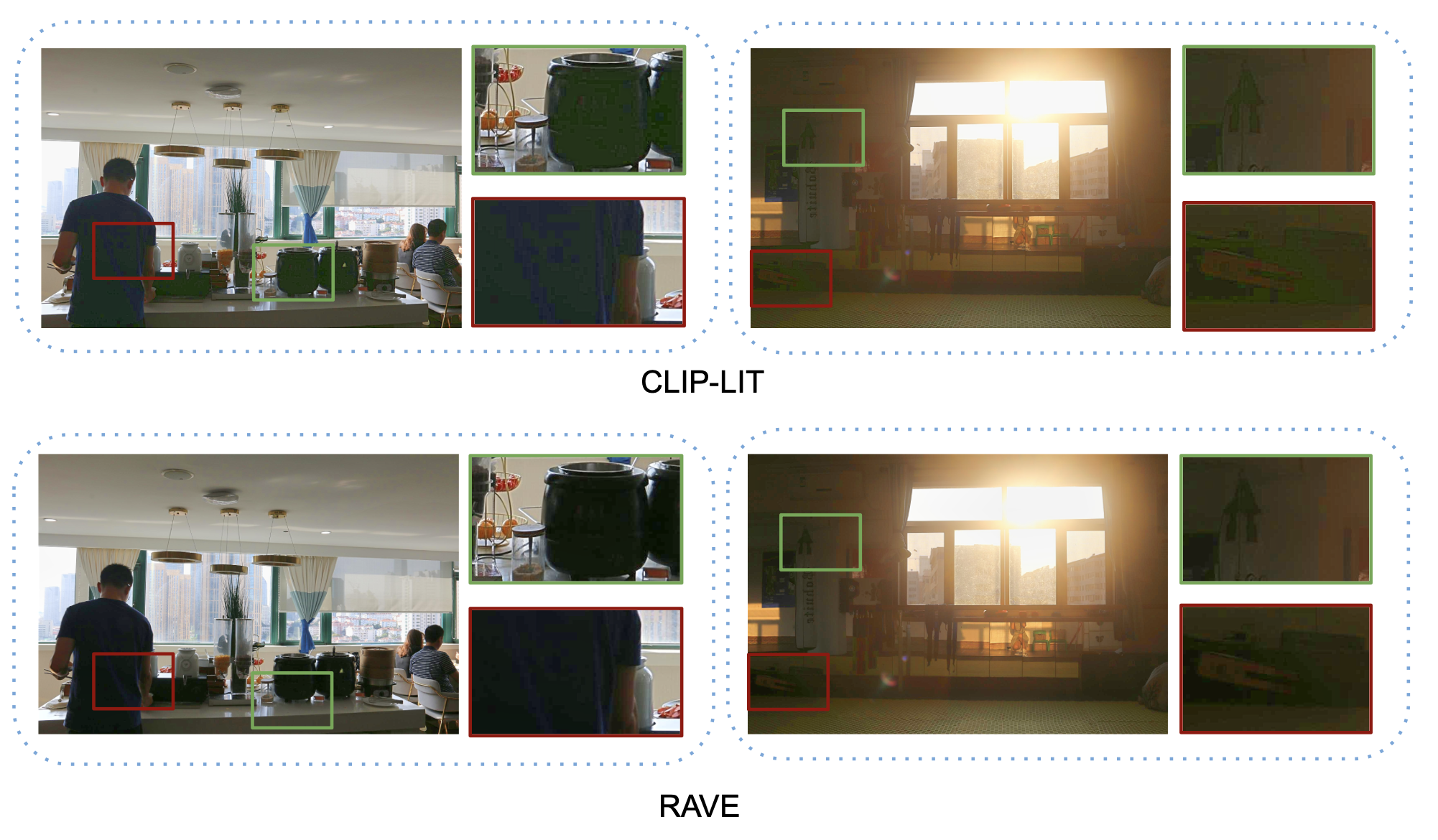}
  \caption{Visual comparison of results obtained by RAVE vs CLIP-LIT. RAVE produces well-lit images with fewer artifacts (note the dark green color of underexposed regions in CLIP-LIT), as shown in the zoomed-in sections. Further comparisons can be found in supplementary material.
  }
  \label{fig:artifacts}
\end{figure}
Backlit image enhancement aims to improve images that suffer from backlighting issues. Backlighting is a phenomenon when the light source is situated behind the photographed object, often resulting in a loss of detail and contrast of some areas due to underexposure, diminishing the overall visual quality of the image. The goal of backlit image enhancement is to correct these exposure discrepancies and bring the details in the darkened areas out without overexposing the well-exposed parts of the image. Backlit image enhancement significantly improves photograph quality in uncontrolled lighting conditions. It is also vital for fields requiring clear image visibility, such as surveillance, security, and scientific imaging applications.

However, correcting backlit images is not an easy task. Manual correction requires skill using photo enhancement software, and often substantial time and effort. Automated solutions are preferred, but also face challenges. Global adjustment of the brightness level is not sufficient, as this typically overexposes regions of the image which were well lit prior to enhancement \cite{9609683}. Spatially adaptive methods have appeared in the literature including early methods that rely on assumptions such as average luminance and a gray world model \cite{9157813, 9369102} or Retinex theory \cite{land1977retinex}. More recent methods approximate well-lit image distributions via end-to-end training of deep networks \cite{restormer, retinexformer}. Despite progress in the field, considerable room for improvement exists for high quality automated backlit image enhancement.

Another challenge to high quality automatic backlit image enhancement is the lack of paired data, i.e. backlit images each with a corresponding well-lit image. It is complicated to collect such data, so approaches that work with unpaired data are imperative. Recently, CLIP-LIT\cite{clip-lit} was proposed, which successfully utilizes CLIP\cite{Radford2021LearningTV} model guidance for training a backlit enhancement model with unpaired data. CLIP-LIT uses prompt learning techniques to provide CLIP guidance. More specifically, it constructs two learnable text prompts, which are trained to have CLIP embeddings close to the well-lit and backlit images, respectively. These learnt prompts  then guide the image enhancement model during training. Next, several iterations of additional prompt correction followed by enhancement model fine-tuning using the updated prompts are required to produce the final image enhancement model. This approach is the current state-of-the-art in backlit image enhancement with training on unpaired data. 

In this work, we demonstrate that prompt training is not the most efficient way to implement the CLIP guidance, and propose two novel methods, named \emph{CLIP-LIT-Latent} and \emph{ResiduAl Vector Embedding (RAVE)}. In CLIP-LIT-Latent, we train vectors corresponding to well-lit and backlit images directly in the CLIP latent space rather than in text embedding space. We show that this approach has similar or superior performance when trained on paired or unpaired image data. Further, in RAVE, we form the guidance vector by subtracting the mean of CLIP latent vectors of backlit images from the  mean of CLIP latent vectors of well-lit images. This residual vector points in a direction of moving backlit images to well-lit images in the CLIP embedding space. This vector is then used as guidance for the image enhancement model during training. We show that such guidance enables the enhancement model to produce high quality images, while requiring much less training time. While having all these benefits, both proposed CLIP-LIT-Latent and RAVE retain all the advantages of the original CLIP-LIT method, being lightweight for inference and suitable for training on paired or unpaired data. Moreover, we show that residual vector used for guidance in RAVE is interpretable, which opens up new possibilities for overcoming biases in the training data.

\medskip
The contributions of this paper can be summarized as follows:

\begin{enumerate}

    \item Inspired by CLIP-LIT, we present two novel approaches to CLIP guidance (CLIP-LIT-Latent and RAVE) working directly in the latent space for the task of backlit image enhancement. 
    \item We show that these approaches result in similar or better quality according to quantitative metrics in both settings of training on paired or unpaired data, and superior visual quality of resulting images, while requiring considerably less time to train.
    \item We demonstrate that the embedding used by RAVE for guidance is interpretable, and its interpretation can reveal biases in the training data. 
\end{enumerate}

\section{Related work}

\noindent\textbf{Image Enhancement.} Image enhancement is the task of image processing with the goal of improving the visual appearance of an image. This is an essential task in numerous fields, such as photography, medical imaging, satellite imagery analysis, and other applications where image quality is crucial. As there are many types of image degradation such as noise, blurriness, low resolution, backlighting, low light, etc., there are a range of image enhancement methods, from traditional ones such as gray level transformation\cite{9157813, 9369102}, histogram equalization\cite{cheng2004simple}, and methods based on Retinex theory\cite{land1977retinex}, to machine learning and end-to-end trained deep learning methods \cite{restormer, retinexformer}. A general overview of concepts and techniques commonly used for image enhancement is provided by Maini $\etal$\cite{8878706}. 

\noindent\textbf{Backlit Image Enhancement.}
For the task of backlit image enhancement, many approaches have been proposed. Some techniques aim to identify backlit regions and correct them separately from the rest of the image\cite{8101559}. Methods based on the Retinex\cite{land1977retinex} theory divide an image into a reflectance image and an illumination map, and both parts are then enhanced and then combined back together to form the resulting enhanced image. Another approach to enhancing backlit images is using end-to-end deep learning approaches, as in Restormer\cite{restormer} or Retinexformer\cite{retinexformer} methods. Combined techniques based on Retinex theory and deep learning were proposed for both backlit and low-light image enhancement, such as Diff-Retinex\cite{diff-retinex}. These methods decomposing an input image into a reflectance image and an illumination map and apply diffusion methods for enhancement before combining them to produce the output image. Finally, another set of methods are based on high dynamic range (HDR) imaging which combine multiple exposures to produce a composite well-lit image\cite{Reinhard2020, hdri, CatleyChandhar2022}.  However, in this work we only focus on single-image methods where back-lit correction is applied as a post-process. 

Fundamentally, backlit image correction is ill-posed, in that for a single backlit image, multiple well-lit variants with high perceptual quality exist. Relatedly, generation of paired training data requires hiring professional photographers, which can be expensive. To help address this problem, recently the BAID\cite{baid} dataset was proposed, which contains paired images generated by professional photographers. Nevertheless, development of methods for backlit image enhancement, which can be trained using non-paired data remains essential. 

\noindent\textbf{CLIP guidance.}
CLIP\cite{Radford2021LearningTV} is a model that consists of two parts: image and text encoders, which learn to project images and text to a common latent embedding space. CLIP has been proven to provide strong image and text priors which are able to provide guidance in many tasks. CLIP guidance has been used in text-to-image generation\cite{glide}, visual question answering\cite{10208332}, GAN inversion\cite{baykal2023clipguided}, 3D scene generation\cite{111111}, domain adaptation of image generators\cite{clip-da} and more. Liang $\etal$\cite{clip-lit} also used CLIP guidance for the task of backlit image enhancement. In our work, we aim to improve this technique by modifying the CLIP guidance.

\section{Methodology}

Our work is built on top of the CLIP-LIT\cite{clip-lit} approach, which uses CLIP guidance for training a backlit image enhancement model. We propose two modifications of CLIP guidance, which reduce training time, stabilize training and produce high quality enhanced images with fewer artifacts compared to CLIP-LIT. 

We start in Section 3.1 by reviewing the architecture of original CLIP-LIT method on which our work is based. Then, in Sections 3.2 and 3.3 we introduce our proposed methods CLIP-LIT-Latent and RAVE. 

\subsection{CLIP-LIT overview}

\begin{figure}[tb]
    \begin{subfigure}{12cm}
  \includegraphics[height=4.9cm]{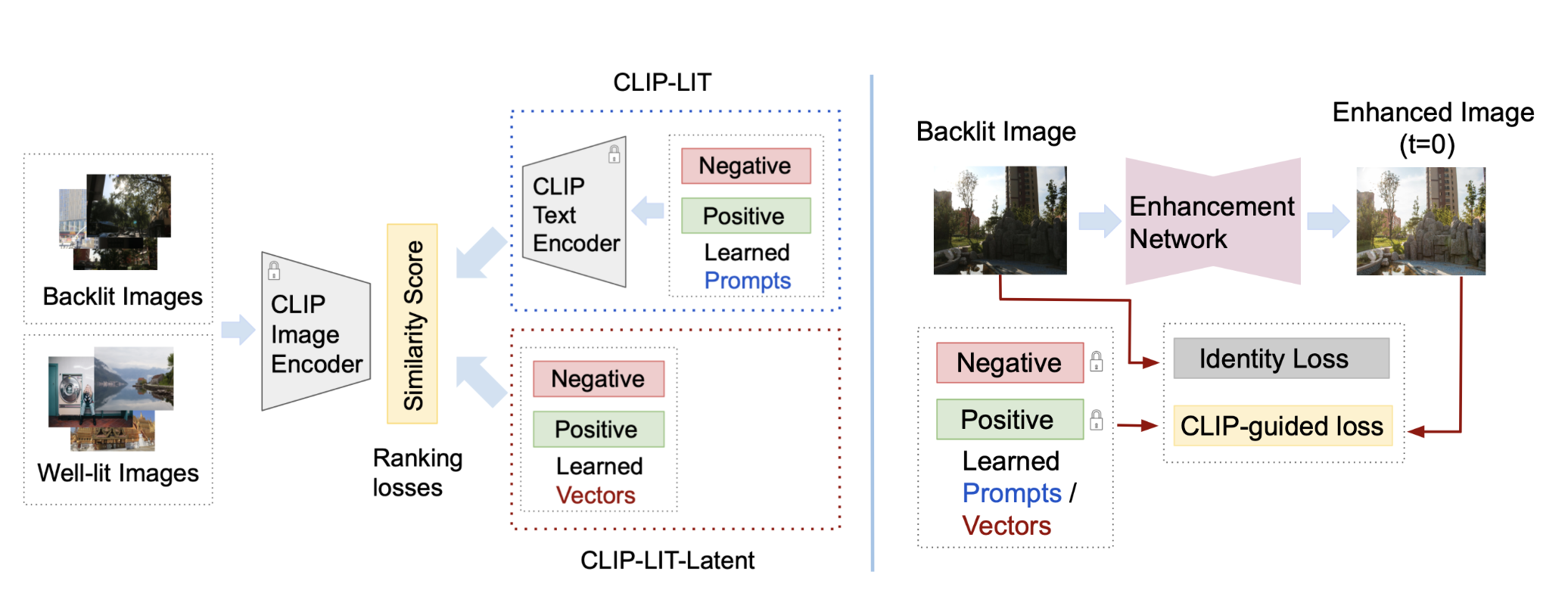}
  \caption{Initial Prompts and Enhancement Training
  }
  \end{subfigure}
  \begin{subfigure}{12cm}
  \includegraphics[height=4.9cm]{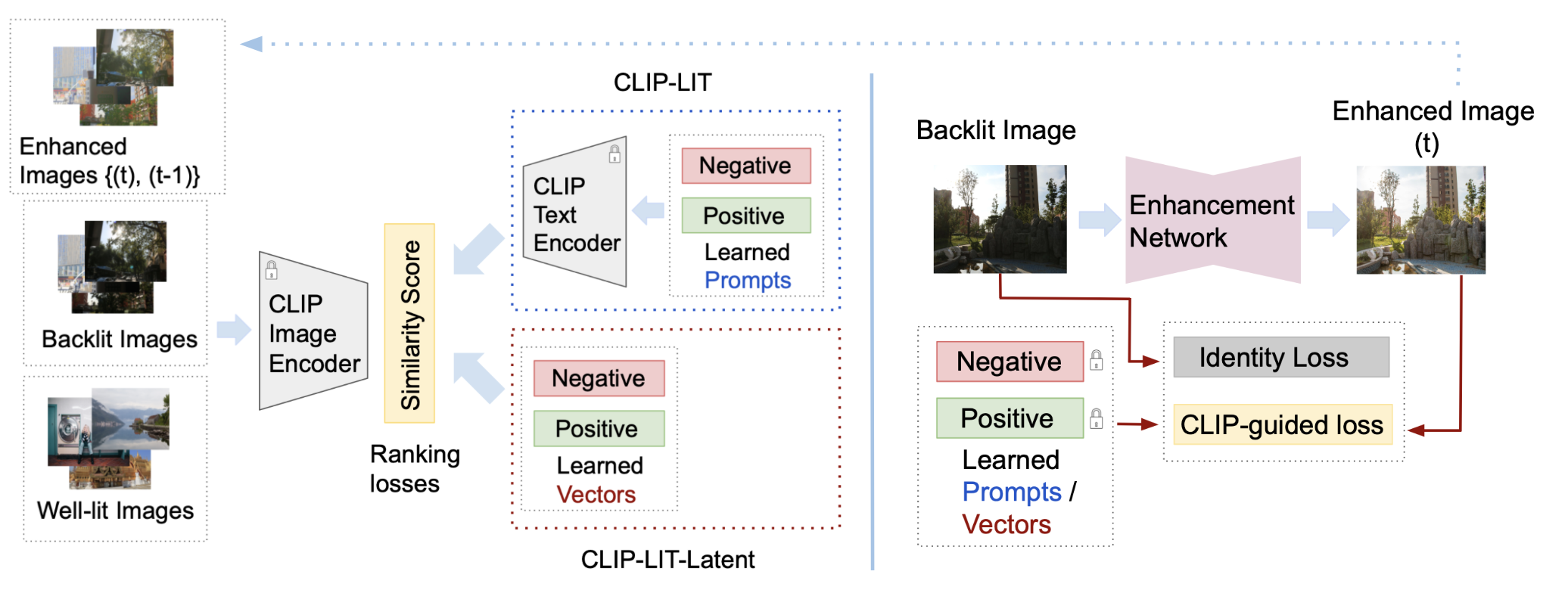}
  \caption{Prompt Refinement and Enhancement Tuning
  }
  \end{subfigure}
  
  \caption{Overview of the original CLIP-LIT approach and proposed CLIP-LIT-Latent. (a) depicts the first stage of training, which consists of prompt or latent vector initialization and the initial training of an enhancement network. (b) shows the second stage, where prompt/latent vector refinement and enhancement model fine-tuning are iteratively repeated. \textcolor{blue}{Blue} and \textcolor{red}{red} boxes are related to CLIP-LIT and CLIP-LIT-Latent.
  }
  \label{fig:clip-lit-orig}
\end{figure}

The CLIP-LIT framework (Fig. ~\ref{fig:clip-lit-orig}) consists of two stages, described below. 
\newline

\noindent\textbf{Initial Prompts and Enhancement Training}
\newline
In this stage, two prompts $T_p, T_n \in \mathbb{R}^{N \times 512}$ representing positive and negative prompts are randomly initialized, where $N$ is a number of tokens in each prompt. Then they are tuned so that the CLIP embedding of $T_p$ is close in terms of the dot product to the CLIP embeddings of well-lit training images, and CLIP embedding of $T_n$ is close to the CLIP embeddings of backlit images from the training data. The binary cross-entropy loss of classifying the backlit and well-lit images based on $T_p$ and $T_n$ is used for this purpose:
\begin{equation}
    \begin{split}
L_{\text{initial}} = -(y \log(\widehat{y}) + (1-y)  \log(1-\widehat{y})) \\
\widehat{y} = \frac{e^{\cos(\Phi_{\text{image}}(I), \Phi_{\text{text}}(T_p))}}{\sum_{i \in\{n, p\}} e^{\cos(\Phi_{\text{image}}(I), \Phi_{\text{text}}(T_i))}}
\end{split}
\end{equation}
Here $I$ is a well-lit or backlit input image with label $y \in \{0, 1\}$ with 0 corresponding to backlit and 1 corresponding to well-lit; $\Phi_{\text{image}}$ is a CLIP\cite{Radford2021LearningTV} image encoder, $\Phi_{\text{text}}$ is a CLIP text encoder; $\cos$ is a cosine similarity function.

Using these initialized prompts $T_p$ and $T_n$ it is now possible to train an image enhancement network with a CLIP-guided loss. The enhancement model is a Unet\cite{unet} which takes backlit image $I_b$ as input and outputs an estimated illumination map $I_i \in \mathbb{R}^{H \times W \times 1}$. The final enhanced image $I_t$ is then obtained as $I_t = I_b / I_i$. The image enhancement network is trained by combining two losses:
\begin{equation}
    L_{\text{enhance}} =  L_{\text{clip}} + \omega  L_{\text{identity}} 
\end{equation}
Here $L_{\text{clip}}$ measures the similarity between the enhanced image and the learned prompts $T_p$ and $T_n$ in the CLIP latent space. $L_{\text{identity}}$ is an identity loss designed to ensure that the enhanced image is similar to the original input image in terms of content and structure. The parameter $\omega$ balances the two losses. These losses are defined as:
\begin{equation}
L_{\text{clip}} = \frac{e^{\cos(\Phi_{\text{image}}(I_t), \Phi_{\text{text}}(T_n))}}{\sum_{i\in\{n, p\}} e^{\cos(\Phi_{\text{image}}(I_t), \Phi_{\text{text}}(T_i))}}
\end{equation}
\begin{equation}
L_{\text{identity}} = \sum_{l=0}^k \alpha_l \cdot \|\Phi^l_{\text{image}}(I_b) - \Phi^l_{\text{image}}(I_t)\|_2
\end{equation}
where $\Phi^l$ is $l^{th}$ layer of the CLIP image encoder, and $\alpha_l$ is its corresponding weight. $k$ is number of layers used to calculate the metric.
\newline

\noindent\textbf{Prompt Refinement and Enhancement Tuning}
\newline
CLIP-LIT applies a second stage where several cycles of prompt refinement and enhancement network tuning are performed to improve the accuracy of the learned prompts.

First, learned prompts $T_p$ and $T_n$ are fine-tuned to provide better guidance for fine-tuning the image enhancement model. A margin ranking loss is used to update the prompts based on images given by the current enhancement model:
\begin{equation}
\begin{split}
    L^1_{\text{prompt}} &= \max(0, S(I_w) - S(I_b) + m_0) + \max(0, S(I_{t-1}) - S(I_b) + m_0) \\
& + \max(0, S(I_w) - S(I_t) + m_1) + \max(0, S(I_t) - S(I_{t-1}) + m_2)
\end{split}
\end{equation}
where $S(I)$ is a negative similarity score between the prompt pair and an image defined as:
\begin{equation}
S(I) = \frac{e^{\cos(\Phi_{\text{image}}(I), \Phi_{\text{text}}(T_n))}}{\sum_{i\in\{n, p\}} e^{\cos(\Phi_{\text{image}}(I), \Phi_{\text{text}}(T_i))}}
\end{equation}
Here $m_0 \in [0, 1]$ is the margin between the score of well-lit or enhanced result and the backlit image in the CLIP embedding space; $m_1 \in [0, 1]$ is the margin between the score of the enhanced result and the well-lit image in the CLIP embedding space; $m_2 \in [0, 1]$ is the margin between the newly enhanced result $I_t$ and previously enhanced result $I_{t-1}$.

After that, a stage of enhancement model fine-tuning is performed. This follows the same process as initial enhancement model training described above. The process of prompt refinement and enhancement tuning is repeated several times until satisfactory results are obtained.

\subsection{CLIP-LIT-Latent}
\label{sect:CLIP-LIT-Latent}

Like CLIP-LIT, CLIP-LIT-Latent also consists of two stages of initial prompts and enhancement training and prompt refinement and enhancement tuning. Fig. ~\ref{fig:clip-lit-orig} illustrates the CLIP-LIT-Latent framework. 

Instead of learning prompts in the text embedding space, CLIP-LIT-Latent learns a pair of positive/negative vectors directly in the CLIP latent space. In the prompt initialization phase, we randomly initialize two vectors $E_p, E_n \in \mathbb{R}^{512}$. These vectors are then tuned so that $E_p$ is close in terms of the dot product to the CLIP embedding of well-lit images from the training data, and $E_n$ is close to the CLIP embedding of backlit images. As in original CLIP-LIT approach, binary cross-entropy loss for classifying the backlit and well-lit images based on $E_p$ and $E_n$ is used for this purpose:
\begin{equation}
    \begin{split}
L_{\text{initial}} = -(y \log(\widehat{y}) + (1-y)  \log(1-\widehat{y}))\\
\widehat{y} = \frac{e^{\cos(\Phi_{\text{image}}(I), E_p)}}{\sum_{i\in\{n, p\}} e^{\cos(\Phi_{\text{image}}(I), E_i)}}
\end{split}
\end{equation}
where $I$ is a well-lit or backlit input image with label $y \in {0, 1}$ with 0 corresponding to backlit and 1 corresponding to well-lit; $\Phi_{\text{image}}$ is a CLIP image encoder. 

Based on these initialized vectors $E_p$ and $E_n$, an enhancement network is trained with a CLIP-guided loss. As before, the enhancement model is a Unet which takes backlit image $I_b$ as input and outputs an estimated illumination map $I_i \in \mathbb{R}^{H \times W \times 1}$. And, as before, it is trained using $L_{\text{enhance}}$ loss, where the $L_{\text{clip}}$ component is changed as follows:
\begin{equation}
L_{\text{clip}} = \frac{e^{\cos(\Phi_{\text{image}}(I_r), E_n)}}{\sum_{i\in\{n, p\}} e^{\cos(\Phi_{\text{image}}(I_t), E_i)}}
\end{equation}
where $\Phi^l$ is $l^{th}$ layer of CLIP image encoder, and $a_l$ is its corresponding weight. 

In the second stage of prompt refinement and enhancement tuning, where learned vectors $E_p$ and $E_n$ are fine-tuned using margin ranking loss, the formula for $S(I)$ changes as follows:
\begin{equation}
S(I) = \frac{e^{\cos(\Phi_{\text{image}}(I), E_n)}}{\sum_{i\in\{n, p\}} e^{\cos(\Phi_{\text{image}}(I), E_i)}}
\end{equation}
In all other aspects the training algorithm is similar to that of CLIP-LIT.

\medskip
CLIP-LIT-Latent has several advantages over CLIP-LIT. First, CLIP-LIT-Latent does not use the text encoder, which speeds up the training and inference processes, as it is no longer necessary to pass vectors and gradients through the text encoder. This also potentially enables the use of other vision models which do not have text encoder parts rather than CLIP for guidance. Second, images obtained from CLIP-LIT-Latent tend to have more contrast and have better visual quality (see Fig. \ref{fig:vis_comp} and Fig. \ref{fig:artifacts}). We speculate this results from tuning vectors $E_p$ and $E_n$ directly in the CLIP latent space. In contrast, in the original CLIP-LIT, vectors $T_p$ and $T_n$ belong to the text embedding space, and must be tuned so that their projection through the CLIP text encoder satisfies desired conditions. Such optimization is more complex and makes tuning of $T_p$ and $T_n$ harder and less accurate. 
\subsection{RAVE: Residual Vector Embedding}
\label{sect:RAVE}

\begin{figure}[tb]
    \begin{subfigure}{.55\textwidth}
  \includegraphics[width=1.0\linewidth]{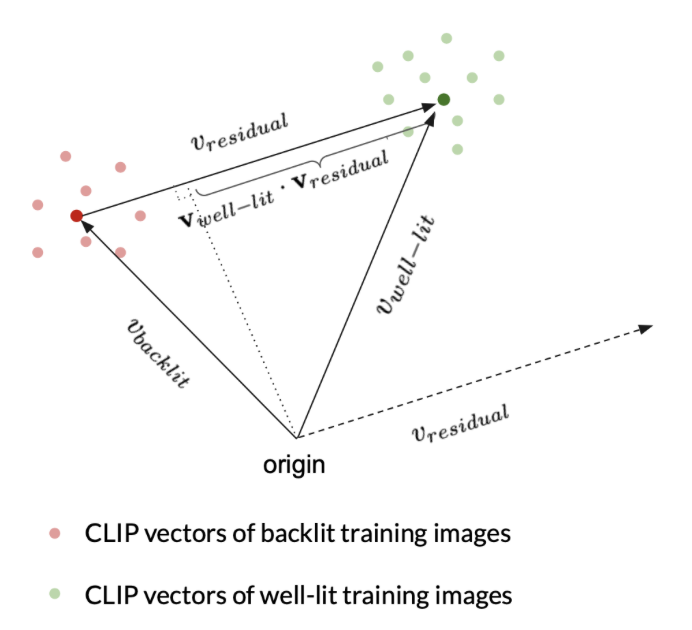}
  \caption{Calculation of residual vector $\mathbf{v}_{\textit{residual}}$ based on paired or unpaired training data }
  \end{subfigure}
  \begin{subfigure}{.45\textwidth}
  \includegraphics[width=1\linewidth]{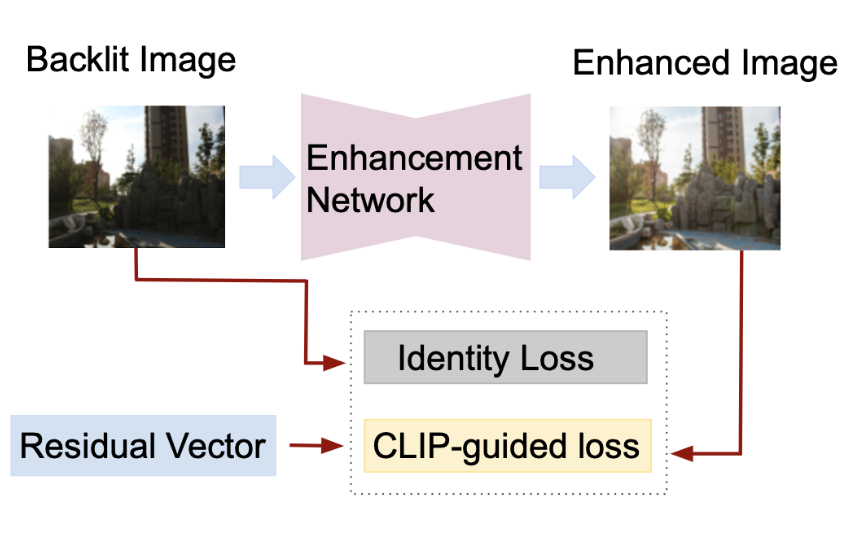}
  \caption{Training of enhancement model using guidance from $\mathbf{v}_{\textit{residual}}$}
  \end{subfigure}
  \caption{Overview of the RAVE model. (a) First, we calculate residual vector $\mathbf{v}_{\text{residual}}$ based on backlit and well-lit training data. (b) Then we switch to enhancement model training based on the identity loss and the loss based on the residual vector.
  }
  \label{fig:rave}
\end{figure}
In this section we present our model RAVE (ResiduAl Vector Embedding).  RAVE includes a further modification of the CLIP guidance. We show that RAVE does not require iterative stages of prompt and model updates.  Instead it only requires \emph{one stage} of enhancement model training.

In RAVE we exploit arithmetic defined in the CLIP latent space. Using well-lit and backlit training data, we construct a residual vector $\mathbf{v}_{\text{residual}}$, which will then be used for enhancement model guidance. $\mathbf{v}_{\text{residual}}$ is defined as follows:
\begin{equation}
\label{eq:12}
    \begin{split}
        & \mathbf{v}_{\text{residual}} =  f_{\text{norm}}(\mathbf{v}_{\text{well-lit}} - \mathbf{v}_{\text{backlit}}) \\
        & \mathbf{v}_{\text{well-lit}} = f_{norm}\left( \frac{\sum_{i=1}^{N_p} f_{\text{norm}}(\Phi_{\text{image}}(I^p_i))}{N_p}\right) \\
        & \mathbf{v}_{\text{backlit}} = f_{norm}\left( \frac{\sum_{i=1}^{N_n} f_{\text{norm}}(\Phi_{\text{image}}(I^n_i))}{N_n} \right)
    \end{split}
\end{equation}
where $f_{\text{norm}}$ is a normalization function: $f_{\text{norm}}(\mathbf{v}) = \frac{\mathbf{v}}{\|\mathbf{v}\|_2}$, $N_p$ is the number of training well-lit images, $N_n$ is the number of training backlit images, $I^p_i$ is the $i^{th}$ well-lit training image, $I^n_i$ is the $i^{th}$ backlit training image.

$\mathbf{v}_{\text{residual}}$ is a vector that points in a direction moving from backlit images to well-lit images in the CLIP embedding space. We then use this vector as guidance for the image enhancement model during training. This will train the image enhancement model to produce images with CLIP latent vectors that are close to the CLIP latent vectors of well-lit training images. Overview of the RAVE approach is presented on Fig. \ref{fig:rave}.

We use the same Unet\cite{unet} as enhancement model as in CLIP-LIT and CLIP-LIT-Latent. The enhancement model is trained using the following loss function:
\begin{equation}
L_{\text{rave}} = L_{\text{identity}} + \omega L_{\text{residual}}
\end{equation}
\begin{equation}
\label{eq:14}
L_{\text{residual}} = || \Phi_{\text{image}}(I) \cdot \mathbf{v}_{\text{residual}} - \mathbf{v}_{\text{well-lit}} \cdot \mathbf{v}_{\text{residual}}||^2
\end{equation}
where $I$ is an input image, and $v_{\text{well-lit}}$ is a normalized mean of CLIP embeddings of well-lit training images:
\begin{equation}
\mathbf{v}_{\text{well-lit}} = f_{\text{norm}} \left( \frac{\sum_{i=1}^{N_p} f_{\text{norm}}(\Phi_{\text{image}}(I^p_i))}{N_p} \right)
\end{equation}

Note that RAVE does not require several repeating stages during training. Before training, the residual vector $\mathbf{v}_{\text{residual}}$ is calculated, and afterwards the training of the image enhancement model begins using $L_{\textit{rave}}$ as the loss function. This makes RAVE training much more efficient than that of CLIP-LIT and CLIP-LIT-Latent. First, the training of the image enhancement model no longer depends on the quality of learned prompt embeddings, as a fixed residual vector is used for guidance. Also, the iterative manner of CLIP-LIT and CLIP-LIT-Latent training regimes makes training less stable, as the image enhancement model training depends on the quality of prompts obtained on the previous step and vice versa. In supplementary material we show that results obtained on intermediate steps of CLIP-LIT training can be of poor quality and can fluctuate from producing under-exposed to over-exposed resulting images from step to step. In contrast, RAVE has a well-defined fixed initial objective. Results show that RAVE converges up to 25 times faster than CLIP-LIT and CLIP-LIT-Latent. Furthermore, images obtained by RAVE have fewer over-exposed areas, producing high quality results.

\section{Experiments}
\noindent\textbf{Datasets}
Following CLIP-LIT, we use the BAID\cite{baid} and DIV2K\cite{div2k} datasets for training. We train our models in two settings: using paired and unpaired training data. For the paired setting, we use 380 backlit and corresponding well-lit images from BAID dataset. For the unpaied setting, we use 380 backlit images from the BAID and 384 images from DIV2K dataset. For the unpaired setting, we use the same data as used in CLIP-LIT. 
For testing, we follow recent literature and use the BAID\cite{baid} test dataset, which has 368 backlit and corresponding well-lit images. The data is publicly available.

\noindent\textbf{Training and inference}
We follow parameter settings of the original CLIP-LIT approach when setting parameters for CLIP-LIT-Latent and RAVE (where applicable), as well as for reproducing CLIP-LIT. Specific details of training and values of hyperparameters are presented in supplementary material.

\noindent\textbf{Metrics}
Following recent works in image enhancement, we report four metrics: PSNR\cite{5596999}, SSIM\cite{ssim}, and LPIPS\cite{lpips} (AlexNet version) and FID\cite{pmlr-v119-naeem20a}.

\noindent\textbf{Methods for comparison}
The main model which we compare our approaches to is CLIP-LIT\cite{clip-lit}, as it showed superior performance on all mentioned metrics, except FID, to many prior image enhancement models. We also include recently published methods such as Restormer\cite{restormer}, Retinexformer\cite{retinexformer}, DiffIR\cite{diffir} and Diff-Retinex\cite{diff-retinex} to the comparison. 

\subsection{Results}
\textbf{Quantitative comparison}
The quantitative comparison on the BAID test dataset is presented in Tab. \ref{tab:sup_unsup_results}. We report test set metrics of the best model checkpoint obtained during training for all the models. Detailed metrics values for all the checkpoints after each training epoch can be found in supplementary material. Note that many approaches can only be trained using paired data, that is why their results are only present in the ``paired'' section. We see that our method achieves state-of-the-art performance in both settings. In supplementary material, we also present results on time needed to train all the models, and confirm that both CLIP-LIT-Latent and RAVE are much more efficient. 
Note that in the setting of unpaired data we have two versions of RAVE, referred as \textit{RAVE} and \textit{RAVE shifted} in Tab. \ref{tab:sup_unsup_results}. \textit{RAVE shifted} is a variant of RAVE with a simple modification of residual vector, see ``Interpretation of the residual vector'' section below for more details.

\begin{table*}[tb]
  \caption{Quantitative comparison of different methods on the BAID test dataset. The best and second best performances in both settings are \textbf{in bold} and \underline{underlined}.
  }
  \label{tab:sup_unsup_results}
  \centering
  \begin{tabular}{@{}L{2cm}L{3.7cm}|C{1.4cm}C{1.4cm}C{1.4cm}C{1.4cm}C{1.4cm}@{}}%
    \toprule
    
    & Method & PSNR $\uparrow$ & SSIM $\uparrow$ & LPIPS $\downarrow$ & FID $\downarrow$ \\
    \hline
    & Original images & 16.91 & 0.76 & 0.21 & 52.47 \\
    \hline
    \multirow{7}{*}{Paired} & Restormer  & 21.07 & 0.832 & 0.192 & 41.17\\
    & Retinexformer &  22.03 & 0.862 & 0.173 & 45.27\\
    & DiffIR &  21.10 & 0.835 & 0.175 & 40.35\\
    & Diff-Retinex & 22.07 & 0.861 & 0.160 & \underline{38.07}\\
    & CLIP-LIT & \underline{21.93} & 0.875 & 0.156 & 41.16 \\
    & CLIP-LIT-Latent (ours) & 21.84 & \underline{0.877} & \underline{0.155} & 42.56 \\
    & RAVE (ours) & \bf{22.26} & \bf{0.880} & \bf{0.139} & \bf{36.01} \\
    \hline
    \multirow{4}{*}{Unpaired} & EnlightenGAN & 17.55 & 0.864 & 0.196 & 43.50 \\
    & CLIP-LIT & \bf{21.59} & 0.874 & 0.160 & 46.49 \\
    & CLIP-LIT-Latent (ours) & \underline{21.45} & \underline{0.877} & 0.156 & 46.20 \\
    &  RAVE (ours) & 20.39 & 0.861 & \underline{0.155} & \bf{40.11} \\
    &  RAVE shifted (ours) & 21.37 & \bf{0.877} & \bf{0.149} & \underline{40.90} \\
  \bottomrule
  \end{tabular}
\end{table*}
    
\begin{figure}[tb]
\begin{subfigure}{12cm}
\centering
      \includegraphics[width=0.77\textwidth]{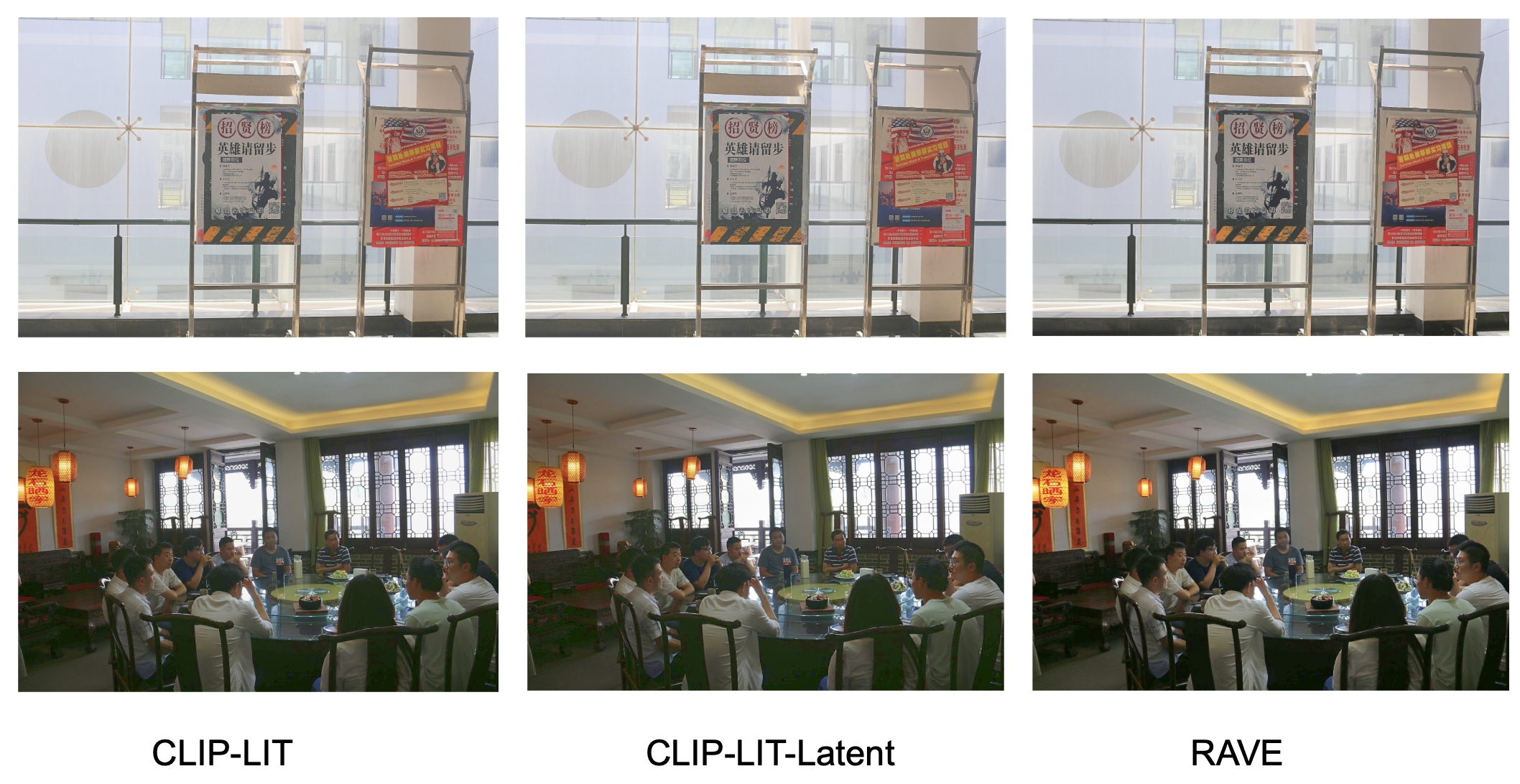}
  \caption{Visual comparison of results by CLIP-LIT, CLIP-LIT-Latent and RAVE, all trained with \emph{paired} data. RAVE produces results with more general contrast.
  }
  \end{subfigure}
\begin{subfigure}{12cm}
\centering
      \includegraphics[width=0.77\textwidth]{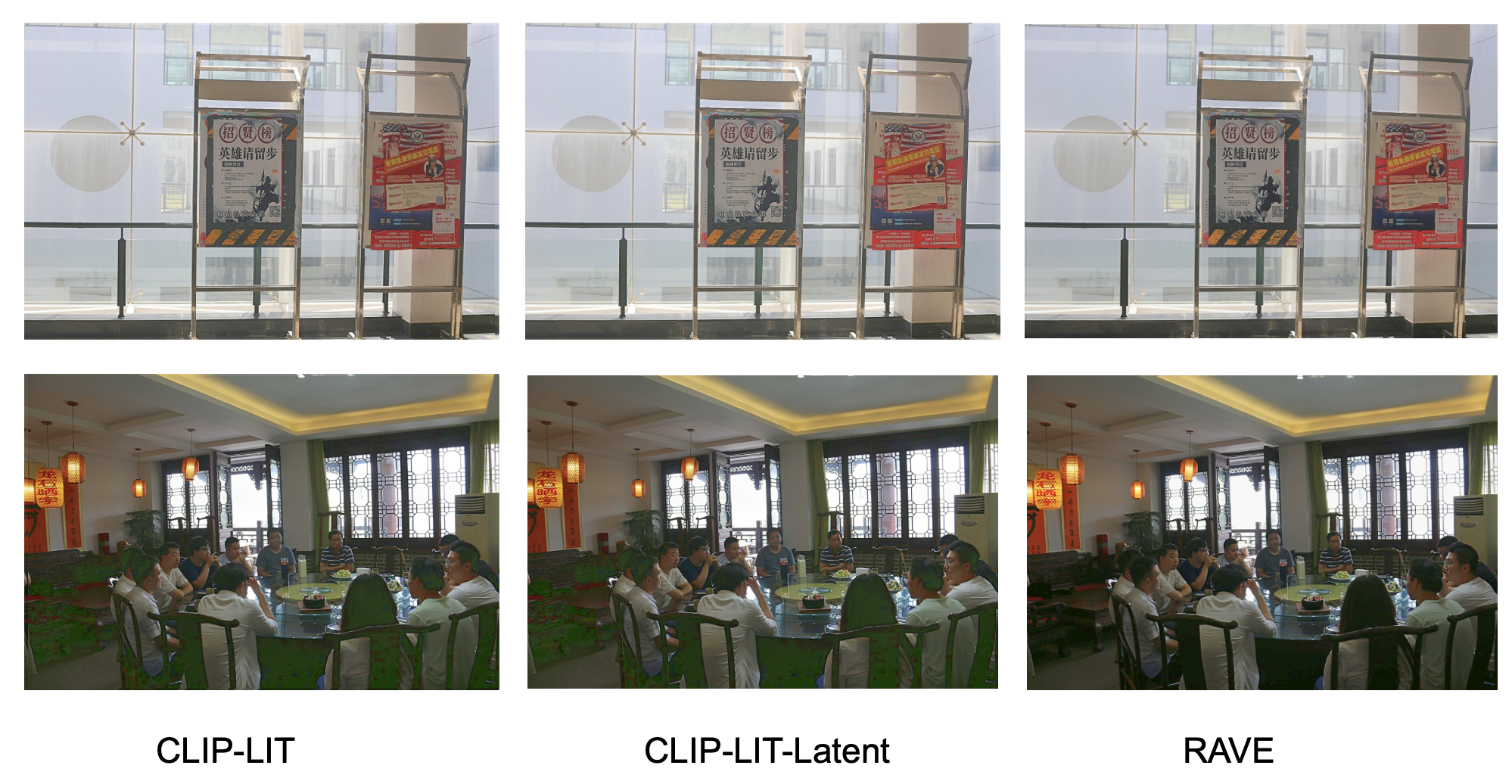}
  \caption{Visual comparison of results obtained by CLIP-LIT, CLIP-LIT-Latent and RAVE (shifted version), all trained with \emph{unpaired} data. RAVE produces results with more general contrast.
  }
  \end{subfigure}
  \caption{Visual comparison of results by CLIP-LIT, CLIP-LIT-Latent and RAVE.
  \label{fig:vis_comp}
  }
\end{figure}
\noindent\textbf{Visual comparison}
For visual comparison, we present representative examples from the BAID test dataset. First, in Fig. \ref{fig:vis_comp}, we compare CLIP-LIT-Latent and RAVE to original CLIP-LIT. First, note that CLIP-LIT-Latent produces more contrast in general, giving more diverse illumination to different regions. Second, in Fig. \ref{fig:artifacts}, we present comparison of images produced by RAVE to images produced by CLIP-LIT. We show that images produced by RAVE have fewer artifacts. We hypothesize that this advantage of RAVE may result from its guiding mechanism: in RAVE we directly point the enhancement model to produce images which have similar CLIP embeddings to that of real well-lit images. This does not allow model to produce regions with artifacts, as its CLIP embeddings won't match those of real well-lit images. In CLIP-LIT or CLIP-LIT-Latent, on the other hand, the objective of the enhancement model is to produce images for which the CLIP embeddings are simply closer to the positive prompt/vector than to the negative one. This allows the enhancement model to produce less accurate images, CLIP embeddings of which are just ``close enough'' to the CLIP embeddings of real well-lit images. More results of visual comparison of all the models can be found in supplementary material.

\noindent\textbf{Interpretation of the residual vector}
\label{sect:interp}
In RAVE, we build the residual vector $\mathbf{v}_{residual}$ following formula \ref{eq:12}. We noted that $\mathbf{v}_{residual}$ is a vector that points to a direction of moving from backlit images to well-lit images in the CLIP embedding space. Here we prove this by interpreting vector $\mathbf{v}_{residual}$ using the fact that CLIP has a common vector space for embeddings of images and text.

We go through the vocabulary of the CLIP text encoder and find those vocabulary tokens which have the closest and farthest CLIP embeddings to the $\mathbf{v}_{residual}$ in the CLIP latent space in terms of cosine similarity. More formally, we use the following metric:
\begin{equation}
    \begin{split}
        sim(token) = f_{\text{norm}}(\Phi_{\text{text}}(e_{\text{token}})) \cdot \mathbf{v}_{\text{residual}}
    \end{split}
\end{equation}
Here $e_{token}$ is embedding of vocabulary $token$, which is input to the CLIP text encoder $\Phi_{text}$. We sort vocabulary tokens using $sim$. Table \ref{tab:emb_sims} shows 7 vocabulary tokens which have highest and lowest cosine similarity to the $v_{residual}$ vectors obtained from the paired dataset and from the unpaired dataset. 

In the case of paired data, many of the least similar tokens have a meaning around "darkness": "silhouette", "dark", "shadow", etc., and absolute value of their cosine similarity to $v_{\text{residual}}$ is around $0.12-0.15$. At the same time, most similar tokens do not share any meaningful semantics, and the absolute value of their cosine similarity to $v_{\text{residual}}$ is much smaller, being around $0.01-0.04$. This supports the hypothesis that $v_{\text{residual}}$ has a meaning of shifting from "dark", "shadowy" images to "normal" images, having average level of lightness. $v_{\text{residual}}$ does not carry any other semantic meaning, as it was calculated using paired dataset, where for each backlit image there is its well-lit analogue, and all the semantics present in the BAID dataset is cancelled as a result of subtraction of the mean well-lit and backlit vectors when constructing $v_{\text{residual}}$. On the other hand, in the case of unpaired data, we see that the least similar tokens have a meaning around "asian": "busan", "beijing", etc., and the most similar ones have a meaning around "nature": "wildlife", "southafrica", "countrylife", etc. Herein, an absolute value of cosine similarities for both sets of tokens if quite high, being around $0.08-0.12$. This reflects the bias in the unpaired datasets. Indeed, in the BAID dataset there are mostly images of Chinese cities or people, while in DIV2K dataset there are many pictures of nature. Due to the bias, RAVE training with unpaired data becomes less effective, making the enhancement model shift images not in an ideal direction in the CLIP latent space. Nevertheless, RAVE still shows good performance when trained with unpaired data. 

We also propose a simple modification for correcting the residual vector in the unpaired setting. We simply take $n$ most and least similar tokens to form an additional residual vector as follows:

\begin{equation}
\label{eq:12}
    \begin{split}
        & \mathbf{v}_{\text{add\_residual}} =  f_{\text{norm}}(\mathbf{v}_{\text{closest}} - \mathbf{v}_{\text{farthest}}) \\
        & \mathbf{v}_{\text{closest}} = f_{norm}\left( \frac{\sum_{i=1}^{n} f_{\text{norm}}(\Phi_{\text{text}}(T^c_i))}{n}\right) \\
        & \mathbf{v}_{\text{farthest}} = f_{norm}\left( \frac{\sum_{i=1}^{n} f_{\text{norm}}(\Phi_{\text{text}}(T^f_i))}{n} \right)
    \end{split}
\end{equation}
$T^c_i$ and $T^f_i$  are tokens with $i^{th}$ most and least similar embeddings to the $\mathbf{v}_{\text{residual}}$. 
And finally we update the $\mathbf{v}_{\text{residual}}$ as follows:
\begin{equation}
\label{eq:12}
    \mathbf{v}_{\text{residual}} =  \mathbf{v}_{\text{residual}} - cos(\mathbf{v}_{\text{residual}}, \mathbf{v}_{\text{add\_residual}}) \cdot \mathbf{v}_{\text{add\_residual}}
\end{equation}

\noindent Resulting vector $\mathbf{v}_{\text{residual}}$ will keep the meaning of shifting from "dark", images to "normal" images while losing the unwanted semantics. Experimental results show that RAVE trained with such updated $\mathbf{v}_{\text{residual}}$ produces better results (see RAVE shifted in Tab. \ref{tab:sup_unsup_results}). In supplementary material we present visual comparison of results obtained by original and shifted versions of RAVE, where it can be clearly seen that RAVE shifted produces more well-lit results, while keeping all the advantages of RAVE. There we also present analysis of tokens with the most and least similar embeddings to the updated $\mathbf{v}_{\text{residual}}$. This analysis supports the claim about the meaning of updated $\mathbf{v}_{\text{residual}}$.

\begin{table}[tb]
  \caption{Vocabulary tokens and their cosine similarities, which have the lowest and highest cosine similarity to the $v_{\text{residual}}$ vectors calculated for paired and unpaired training datasets.}
  \label{tab:emb_sims}
  \centering
  \begin{tabular}{@{}ll|ll|ll|ll@{}}
    \toprule
    \multicolumn{4}{c|}{Paired} & \multicolumn{4}{c}{Unpaired}\\
    \hline
    \multicolumn{2}{c|}{Lowest similarity} & \multicolumn{2}{c|}{Highest similarity} & \multicolumn{2}{c|}{Lowest similarity} & \multicolumn{2}{c}{Highest similarity} \\
    \hline
    silhouette & -0.151 & healthtech & 0.035 & busan & -0.108 & wildlifewednesday & 0.116 \\ 
    darkness & -0.149 & whitepaper & 0.035 & beijing & -0.103 & southafrica & 0.103 \\
    dark  & -0.139 & theeconomist & 0.030 & iu & -0.100 & ecuador & 0.095 \\
    webcamtoy & -0.139 & bona & 0.029 & guangzhou & -0.095 & patrol & 0.095\\
    blackand & -0.132 & digitaltransformation & 0.027 & incheon & -0.094 & countryfile & 0.093 \\
    vedere & -0.131 & sua & 0.021 & shenzhen & -0.094 & tasmanian & 0.093\\
    shadow & -0.130 & amarketing & 0.021 & eunhyuk & -0.091 & womensart & 0.091 \\
  \bottomrule
  \end{tabular}
\end{table}
\noindent\textbf{Necessity of the residual vector}
Given the interpretation of $\mathbf{v}_{\text{residual}}$, we investigate whether it is necessary to compute $\mathbf{v}_{\text{residual}}$ using the training data, or is it enough to use CLIP latent vectors of words such as "dark", "darkness" as a guidance vector for enhancement model. We trained RAVE using $v_{\text{residual}} = \Phi_{\text{text}}(token)$ as guidance vector, with $token \in \{"silhouette", "dark", $$"darkness"\}$, but did not observe any meaningful enhancement in the resulting images. We suspect that this is due to the fact that latent vectors of backlit and well-lit images have similar cosine similarity to embeddings of these tokens, as presented in Tab \ref{tab:mean_sims}. This suggests that residual vector $\mathbf{v}_{\text{residual}}$ is not simply a combination of latent vectors of some tokens, but it contains more specific information with the meaning of shifting from backlit to well-lit data. 
\begin{table}[tb]
  \caption{Mean similarity of CLIP latent vectors of well-lit and back-lit images from BAID and DIV2K datasets to the CLIP latent vectors of vocabulary tokens}
  \label{tab:mean_sims}
  \centering
  \begin{tabular}{@{}L{2cm}|C{2.2cm}C{2.2cm}C{2.2cm}@{}}
    \toprule
    Token & BAID backlit & BAID well-lit & DIV2K well-lit \\
    \hline
    silhouette & 0.254 & 0.227 & 0.223 \\
    dark & 0.247 & 0.222 & 0.232 \\
    darkness & 0.242 & 0.215 & 0.227 \\
  \bottomrule
  \end{tabular}
\end{table}
\subsection{Limitations}
Despite showing great performance in the task of backlit image enhancement, our models have limitation of non-optimal enhancement of the regions where the loss of information occurs, making restoration ambiguous. Our approach has no generative components and cannot regenerate highly-accurate textures when the information is lost. This leads to sub-optimal performance in such tasks as low-light image enhancement, where images are highly under-exposed (see supplementary material). Moreover, our models are not trained to exactly match pixels of resulting images with ground-truth data, so some sub-optimal lightening of pixels might occur, which results in non-ideal visual quality.

However, existing generative-based models also have limitations. Many of them cannot be trained with unpaired data setting, and require much time and memory to be trained and used. They may also hallucinate detail. As a consequence, the resolution of images it can work with is limited.

\section{Conclusion}
Based on CLIP-LIT method, we have presented two improved CLIP guidance techniques for backlit image enhancement task, named CLIP-LIT-Latent and RAVE. Our methods show similar or superior performance to that of CLIP-LIT in both cases of training with paired or unpaired data, while having lower time and space requirements. Furthermore, latent vector constructed by RAVE is interpretable and can be used to find biases in the training data, as well as potentially correct them. Moreover, our proposed approach has several advantages over the current SOTA in backlit image enhancement, Diff-Retinex. First, unlike Diff-Retinex, our method can also be trained using unpaired data. Second, it is more lightweight, requiring less compute and space for training and inference, suitable for images with larger resolution.

\section*{Acknowledgements}
Funding for this research was provided by a Google-DeepMind PhD Studentship, and work utilised Queen Mary's Andrena HPC facility, supported by QMUL Research-IT. We also sincerely thank Laida Kushnareva for assisting with images and text composition in the paper.

\bibliographystyle{splncs04}
\bibliography{main}

\begin{thebibliography}{10}
\providecommand{\url}[1]{\texttt{#1}}
\providecommand{\urlprefix}{URL }
\providecommand{\doi}[1]{https://doi.org/#1}

\bibitem{div2k}
Agustsson, E., Timofte, R.: Ntire 2017 challenge on single image super-resolution: Dataset and study. In: 2017 IEEE Conference on Computer Vision and Pattern Recognition Workshops (CVPRW). pp. 1122--1131 (2017). \doi{10.1109/CVPRW.2017.150}

\bibitem{baykal2023clipguided}
Baykal, A.C., Anees, A.B., Ceylan, D., Erdem, E., Erdem, A., Yuret, D.: Clip-guided stylegan inversion for text-driven real image editing (2023)

\bibitem{retinexformer}
Cai, Y., Bian, H., Lin, J., Wang, H., Timofte, R., Zhang, Y.: Retinexformer: One-stage retinex-based transformer for low-light image enhancement. In: 2023 IEEE/CVF International Conference on Computer Vision (ICCV). pp. 12470--12479 (2023). \doi{10.1109/ICCV51070.2023.01149}

\bibitem{CatleyChandhar2022}
Catley-Chandar, S., Tanay, T., Vandroux, L., Leonardis, A., Slabaugh, G., Pérez-Pellitero, E.: Flexhdr: Modeling alignment and exposure uncertainties for flexible hdr imaging. IEEE Transactions on Image Processing  \textbf{31},  5923--5935 (2022). \doi{10.1109/TIP.2022.3203562}

\bibitem{cheng2004simple}
Cheng, H.D., Shi, X.: A simple and effective histogram equalization approach to image enhancement. Digital signal processing  \textbf{14}(2),  158--170 (2004)

\bibitem{clip-da}
Gal, R., Patashnik, O., Maron, H., Bermano, A.H., Chechik, G., Cohen-Or, D.: Stylegan-nada: Clip-guided domain adaptation of image generators. ACM Trans. Graph.  \textbf{41}(4) (jul 2022). \doi{10.1145/3528223.3530164}, \url{https://doi.org/10.1145/3528223.3530164}

\bibitem{111111}
Goloujeh, A.M., Smith, J., Magerko, B.: Explainable clip-guided 3d-scene generation in an ai holodeck. In: Proceedings of the Eighteenth AAAI Conference on Artificial Intelligence and Interactive Digital Entertainment. AIIDE'22, AAAI Press (2022). \doi{10.1609/aiide.v18i1.21973}, \url{https://doi.org/10.1609/aiide.v18i1.21973}

\bibitem{9157813}
Guo, C., Li, C., Guo, J., Loy, C.C., Hou, J., Kwong, S., Cong, R.: Zero-reference deep curve estimation for low-light image enhancement. In: 2020 IEEE/CVF Conference on Computer Vision and Pattern Recognition (CVPR). pp. 1777--1786 (2020). \doi{10.1109/CVPR42600.2020.00185}

\bibitem{5596999}
Horé, A., Ziou, D.: Image quality metrics: Psnr vs. ssim. In: 2010 20th International Conference on Pattern Recognition. pp. 2366--2369 (2010). \doi{10.1109/ICPR.2010.579}

\bibitem{land1977retinex}
Land, E.H.: The retinex theory of color vision. Scientific american  \textbf{237}(6),  108--129 (1977)

\bibitem{9609683}
Li, C., Guo, C., Han, L., Jiang, J., Cheng, M., Gu, J., Loy, C.: Low-light image and video enhancement using deep learning: A survey. IEEE Transactions on Pattern Analysis and Machine Intelligence  \textbf{44}(12),  9396--9416 (dec 2022). \doi{10.1109/TPAMI.2021.3126387}

\bibitem{9369102}
Li, C., Guo, C., Loy, C.C.: Learning to enhance low-light image via zero-reference deep curve estimation. IEEE Transactions on Pattern Analysis and Machine Intelligence  \textbf{44}(8),  4225--4238 (2022). \doi{10.1109/TPAMI.2021.3063604}

\bibitem{8101559}
Li, Z., Wu, X.: Learning-based restoration of backlit images. IEEE Transactions on Image Processing  \textbf{27}(2),  976--986 (2018). \doi{10.1109/TIP.2017.2771142}

\bibitem{clip-lit}
Liang, Z., Li, C., Zhou, S., Feng, R., Loy, C.C.: Iterative prompt learning for unsupervised backlit image enhancement. In: 2023 IEEE/CVF International Conference on Computer Vision (ICCV). pp. 8060--8069 (2023). \doi{10.1109/ICCV51070.2023.00743}

\bibitem{baid}
Lv, X., Zhang, S., Liu, Q., Xie, H., Zhong, B., Zhou, H.: Backlitnet: A dataset and network for backlit image enhancement. Computer Vision and Image Understanding  \textbf{218},  103403 (2022). \doi{https://doi.org/10.1016/j.cviu.2022.103403}, \url{https://www.sciencedirect.com/science/article/pii/S1077314222000340}

\bibitem{pmlr-v119-naeem20a}
Naeem, M.F., Oh, S.J., Uh, Y., Choi, Y., Yoo, J.: Reliable fidelity and diversity metrics for generative models. In: III, H.D., Singh, A. (eds.) Proceedings of the 37th International Conference on Machine Learning. Proceedings of Machine Learning Research, vol.~119, pp. 7176--7185. PMLR (13--18 Jul 2020), \url{https://proceedings.mlr.press/v119/naeem20a.html}

\bibitem{glide}
Nichol, A., Dhariwal, P., Ramesh, A., Shyam, P., Mishkin, P., McGrew, B., Sutskever, I., Chen, M.: Glide: Towards photorealistic image generation and editing with text-guided diffusion models (12 2021)

\bibitem{10208332}
Parelli, M., Delitzas, A., Hars, N., Vlassis, G., Anagnostidis, S., Bachmann, G., Hofmann, T.: Clip-guided vision-language pre-training for question answering in 3d scenes. In: 2023 IEEE/CVF Conference on Computer Vision and Pattern Recognition Workshops (CVPRW). pp. 5607--5612 (2023). \doi{10.1109/CVPRW59228.2023.00593}

\bibitem{Radford2021LearningTV}
Radford, A., Kim, J.W., Hallacy, C., Ramesh, A., Goh, G., Agarwal, S., Sastry, G., Askell, A., Mishkin, P., Clark, J., Krueger, G., Sutskever, I.: Learning transferable visual models from natural language supervision. In: International Conference on Machine Learning (2021), \url{https://api.semanticscholar.org/CorpusID:231591445}

\bibitem{Reinhard2020}
Reinhard, E.: High Dynamic Range Imaging, pp.~1--6. Springer International Publishing, Cham (2020). \doi{10.1007/978-3-030-03243-2_843-1}, \url{https://doi.org/10.1007/978-3-030-03243-2_843-1}

\bibitem{unet}
Ronneberger, O., Fischer, P., Brox, T.: U-net: Convolutional networks for biomedical image segmentation. In: Navab, N., Hornegger, J., Wells, W.M., Frangi, A.F. (eds.) Medical Image Computing and Computer-Assisted Intervention -- MICCAI 2015. pp. 234--241. Springer International Publishing, Cham (2015)

\bibitem{8878706}
Singh, K., Seth, A., Sandhu, H.S., Samdani, K.: A comprehensive review of convolutional neural network based image enhancement techniques. In: 2019 IEEE International Conference on System, Computation, Automation and Networking (ICSCAN). pp.~1--6 (2019). \doi{10.1109/ICSCAN.2019.8878706}

\bibitem{ssim}
Wang, Z., Bovik, A., Sheikh, H., Simoncelli, E.: Image quality assessment: from error visibility to structural similarity. IEEE Transactions on Image Processing  \textbf{13}(4),  600--612 (2004). \doi{10.1109/TIP.2003.819861}

\bibitem{diffir}
Xia, B., Zhang, Y., Wang, S., Wang, Y., Wu, X., Tian, Y., Yang, W., Gool, L.V.: Diffir: Efficient diffusion model for image restoration. In: 2023 IEEE/CVF International Conference on Computer Vision (ICCV). pp. 13049--13059. IEEE Computer Society, Los Alamitos, CA, USA (oct 2023). \doi{10.1109/ICCV51070.2023.01204}, \url{https://doi.ieeecomputersociety.org/10.1109/ICCV51070.2023.01204}

\bibitem{diff-retinex}
Yi, X., Xu, H., Zhang, H., Tang, L., Ma, J.: Diff-retinex: Rethinking low-light image enhancement with a generative diffusion model. In: 2023 IEEE/CVF International Conference on Computer Vision (ICCV). pp. 12268--12277 (2023). \doi{10.1109/ICCV51070.2023.01130}

\bibitem{restormer}
Zamir, S.W., Arora, A., Khan, S., Hayat, M., Khan, F.S., Yang, M.: Restormer: Efficient transformer for high-resolution image restoration. In: 2022 IEEE/CVF Conference on Computer Vision and Pattern Recognition (CVPR). pp. 5718--5729 (2022). \doi{10.1109/CVPR52688.2022.00564}

\bibitem{lpips}
Zhang, R., Isola, P., Efros, A.A., Shechtman, E., Wang, O.: The unreasonable effectiveness of deep features as a perceptual metric. In: 2018 IEEE/CVF Conference on Computer Vision and Pattern Recognition. pp. 586--595 (2018). \doi{10.1109/CVPR.2018.00068}

\bibitem{hdri}
Zheng, B., Pan, X., Zhang, H., Zhou, X., Slabaugh, G., Yan, C., Yuan, S.: Domainplus: Cross transform domain learning towards high dynamic range imaging. In: Proceedings of the 30th ACM International Conference on Multimedia. p. 1954–1963. MM '22, Association for Computing Machinery, New York, NY, USA (2022). \doi{10.1145/3503161.3547823}, \url{https://doi.org/10.1145/3503161.3547823}

\end{thebibliography}
\end{document}


\title{Supplementary Material: RAVE: Residual Vector Embedding for CLIP-Guided Backlit Image Enhancement} 

\titlerunning{RAVE: Residual Vector Embedding}

\author{}

\authorrunning{T.~Gaintseva et al.}

\institute{}

\maketitle

\section{Training Details}

Here we give a detailed overview of the training details of our models. 
\medskip

Our methods are implemented using PyTorch using single NVidia A100 40 GB GPU. We follow parameter settings of the original CLIP-LIT approach when setting parameters for CLIP-LIT-Latent and RAVE (where applicable), as well as for reproducing CLIP-LIT. More specifically, we use the Adam optimizer with $\beta_1 = 0.9$ and $\beta_2 = 0.99$ for all the models. We set total number of iterations to 50k for original CLIP-LIT and CLIP-LIT-Latent, and to 10k for RAVE, though all the methods achieve best performance earlier.

For CLIP-LIT and CLIP-LIT-Latent, when starting training of the enhancement model on the first stage, the training schedule is split into two parts.  For the first 1000 iterations, only the identity loss $L_{identity}$ is used to train the Unet to reconstruct the initial backlit image. Afterwards, the full $L_{enhance}$ loss with $\omega=0.9$ is used for training. The learning rate for the prompt initialization/refinement and enhancement network training are set to $5 \cdot 10^{-6}$ and $2 \cdot 10^{-5}$. The batch size for prompt initialization/refinement and enhancement network training is set to 8 and 16, respectively. Margins values are set to $m_0=0.9$, $m_1=m_2=0.2$. For RAVE, the learning rate for enhancement model training is set to $2 \cdot 10^{-5}$, and the batch size is set to 8. After parameter tuning, we set $\omega=6$ in $L_{rave}$. During training of all the models, the input images are resized to 512 $\times$ 512 and augmentations of flip, zoom, and rotation are used.

During inference, following CLIP-LIT, all the images are resized to have a longer size of $2048$ to ensure a fair comparison.

\newpage
\section{Results on Different Model Checkpoints During Training}

In Tables \ref{tab:sup_epochs_results} and \ref{tab:unsup_epochs_results} we provide quantitative and qualitative results of intermediate model checkpoints obtained during training of CLIP-LIT, CLIP-LIT-Latent and RAVE.

\begin{figure}[hbt!]
  \centering
      \includegraphics[width=\textwidth]{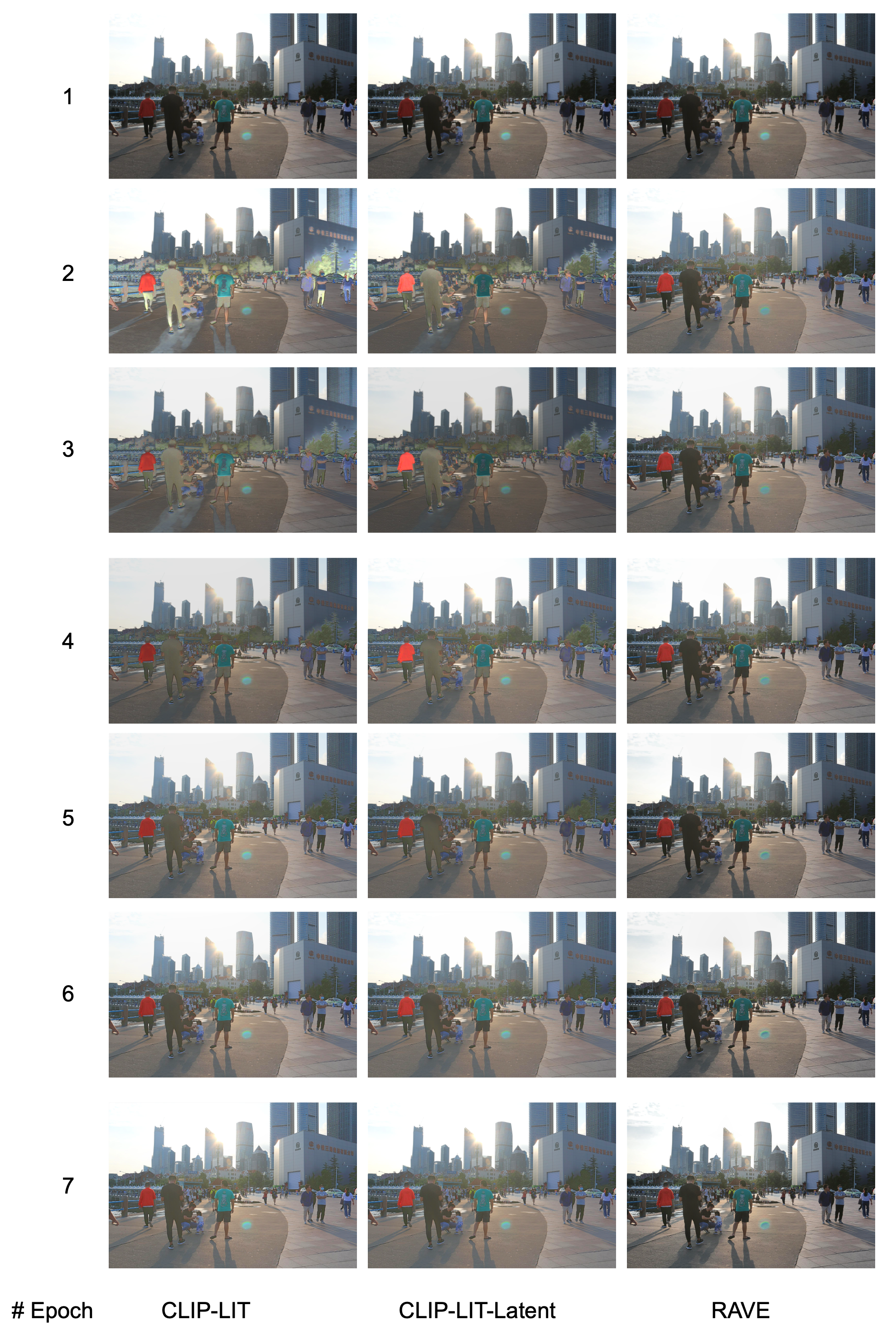}
  \caption{Visualization of performances of CLIP-LIT, CLIP-LIT-Latent and RAVE trained using \textit{unpaired} data on different intermediate checkpoints.
  }
  \label{fig:epochs}
\end{figure}

We also visualize enhanced images obtained from intermediate model checkpoints of CLIP-LIT, CLIP-LIT-Latent and RAVE on Fig. \ref{fig:epochs} trained using \textit{unpaired} data. In both Tab. \ref{tab:unsup_epochs_results} and Fig. \ref{fig:epochs} we see that CLIP-LIT and CLIP-LIT-Latent show decent performance only by the 6th epoch of training, while training of RAVE is much more stable, showing great results starting from 2nd epoch.  

\begin{table*}[hbt!]
  \caption{Quantitative comparison of different checkpoints of different methods on the BAID test dataset using \textit{paired} data. The best and second best performances in all the models are \textbf{in bold} and \underline{underlined}.
  }
  \label{tab:sup_epochs_results}
  \centering
  \begin{tabular}{@{}l|l|l|l|l|l|l|l|l|l|l|l|l|l|l|l@{}}
  
    \toprule
   & \# Epoch & 1 & 2 & 3 & 4 & 5 & 6 & 7 & 8  \\
   \hline
     \multirow{4}{*}{CLIP-LIT} & ($\uparrow$) PSNR & 16.92 & 21.41 & 21.36 & 21.65 & 21.75 & 21.84 & \textbf{21.93} & 21.50  \\
                               & ($\uparrow$) SSIM    & 0.763 & 0.872 & 0.875 & 0.875 & 0.872 & 0.874 & \underline{0.875} & 0.874  \\ 
                               & ($\downarrow$) LPIPS  & 0.211 & 0.161 & 0.160 & 0.158 & 0.165 & 0.158 & \underline{0.157} & 0.161   \\
                               & ($\downarrow$) FID & 52.68 & 41.67 & \underline{41.27} & 41.67 & 46.48 & 42.02 & \textbf{41.17} & 42.80  \\
    \hline
    \multirow{4}{*}{CLIP-LIT-Latent} & ($\uparrow$) PSNR & 16.80 & 20.69 & 21.27 & 20.79 & 21.56 & 20.30 & 21.74 & 21.07      \\
                               & ($\uparrow$) SSIM & 0.763 & 0.855 & 0.864 & 0.863 & 0.870 & 0.866 & 0.876 & 0.875  \\
                               & ($\downarrow$)  LPIPS & 0.213 & 0.189 & 0.178 & 0.187 & 0.170 & 0.186 & 0.161 & 0.163 \\
                               & ($\downarrow$) FID  & 52.42 & 49.71 & 45.22 & 49.92 & 46.07 & 50.84 & \underline{42.47} & \textbf{41.58} \\
    \hline
    \multirow{4}{*}{RAVE} & ($\uparrow$) PSNR  & \underline{22.20} & \textbf{22.26} & 22.13 & 21.78 & 22.00 & 21.28 & 21.93 & 21.96 \\
                               & ($\uparrow$) SSIM & 0.881 & 0.881 & \textbf{0.883} & 0.879 & \underline{0.882} & 0.877 & 0.881 & 0.880 \\
                               & ($\downarrow$) LPIPS & \underline{0.140} & \textbf{0.139} & 0.146 & 0.142 & 0.145 & 0.146 & 0.145 & 0.142 \\
                               & ($\downarrow$) FID & \underline{36.02} & \textbf{36.01} & 38.55 & 36.63 & 39.43 & 37.90 & 40.08 & 38.53 \\
     
  \hline
  \hline

  & \# Epoch & 9 & 10 & 11 \\
   \hline
     \multirow{4}{*}{CLIP-LIT} & ($\uparrow$) PSNR &  \underline{21.83} & 21.36 & 21.60 \\
                               & ($\uparrow$) SSIM    & \textbf{0.876} & 0.870 & 0.875 \\ 
                               & ($\downarrow$) LPIPS  & \textbf{0.157} & 0.169  & 0.158  \\
                               & ($\downarrow$) FID &  41.63 & 47.87 & 41.77 \\
    \hline
    \multirow{4}{*}{CLIP-LIT-Latent} & ($\uparrow$) PSNR & \textbf{21.81} & \underline{21.67} & 21.84     \\
                               & ($\uparrow$) SSIM & \textbf{0.876} & \underline{0.876} & 0.877 \\
                               & ($\downarrow$)  LPIPS & 0.158 & \underline{0.159} & \textbf{0.155} \\
                               & ($\downarrow$) FID  & 43.57 & 43.46 & 42.56 \\
    \hline
    \multirow{4}{*}{RAVE} & ($\uparrow$) PSNR  &  21.79 & 21.89 & 21.90\\
                               & ($\uparrow$) SSIM & 0.877 & 0.879 & 0.879 \\
                               & ($\downarrow$) LPIPS & 0.142 & 0.144 & 0.145\\
                               & ($\downarrow$) FID &  38.07 & 40.18 & 41.56 \\

\bottomrule
  \end{tabular}
\end{table*}

\newpage

\begin{table*}[hbt!]
  \caption{Quantitative comparison of different checkpoints of different methods on the BAID test dataset trained on \textit{unpaired} data. The best and second best performances in all the models are \textbf{in bold} and \underline{underlined}.}
  \label{tab:unsup_epochs_results}
  \centering
  \begin{tabular}{@{}l|l|l|l|l|l|l|l|l|l|l|l|l|l|l|l|l|l@{}}
  
    \toprule
   & \# Epoch & 1 & 2 & 3 & 4 & 5 & 6 & 7 & 8 \\
   \hline
     \multirow{4}{*}{1} & ($\uparrow$) PSNR & 16.93 & 11.60 & 16.15 & 18.13 & 20.83 & 21.20 & 20.92 & 21.31 \\
                               & ($\uparrow$) SSIM & 0.764 & 0.713 & 0.795 & 0.849 & 0.867 & 0.872 & 0.866 & 0.873 \\ 
                               & ($\downarrow$) LPIPS & 0.211 & 0.310 & 0.252 & 0.194 & 0.169 & 0.162 & 0.171 & \underline{0.160}  \\
                               & ($\downarrow$) FID & 52.76 & 127.10 & 99.58 & 64.02 & 56.88 & 51.16 & 57.08 & 47.04\\
    \hline
    \multirow{4}{*}{2} & ($\uparrow$) PSNR & 16.93 & 14.52 & 14.95 & 19.50 & 19.79 & 20.88 & 20.96 & 18.28     \\
                               & ($\uparrow$) SSIM & 0.764 & 0.748 & 0.780 & 0.845 & 0.859 & 0.871 & 0.871 & 0.864  \\
                               & ($\downarrow$)  LPIPS & 0.211 & 0.281 & 0.272 & 0.196 & 0.184 & 0.168 & 0.168 & 0.189  \\
                               & ($\downarrow$) FID & 53.09 & 114.51 & 98.56 & 75.36 & 65.56 & 54.64 & 54.46 & 50.38  \\
    \hline
    \multirow{4}{*}{3} & ($\uparrow$) PSNR & 17.14 & \textbf{20.39} & 19.70 & \underline{19.77} & 18.52 & 18.31 & 18.75 & 18.15 \\
                               & ($\uparrow$) SSIM & 0.769 & \underline{0.861} & \textbf{0.863} & 0.860 & 0.827 & 0.812 & 0.805 & 0.806 \\
                               & ($\downarrow$) LPIPS & 0.207 & \textbf{0.155} & \underline{0.162} & 0.162 & 0.189 & 0.216 & 0.219 & 0.227 \\
                               & ($\downarrow$) FID & 51.74 & \textbf{40.11} & \underline{44.12} & 44.58 & 46.82 & 52.21 & 54.64 & 56.92 \\
     
  \hline
  \hline

  & \# Epoch & 9 & 10 & 11 & 12 & 13\\
   \hline
     \multirow{4}{*}{CLIP-LIT} & ($\uparrow$) PSNR &  21.04 & \underline{21.32} & 17.19 & \textbf{21.60} & 19.55  \\
                               & ($\uparrow$) SSIM & 0.871 & \underline{0.874} & 0.855 & \textbf{0.874} & 0.867\\ 
                               & ($\downarrow$) LPIPS & 0.165 & 0.161 & 0.204 & \textbf{0.160} & 0.171 \\
                               & ($\downarrow$) FID & 50.61 & 47.25 & 52.36 & \underline{46.50} & \textbf{44.44} \\
    \hline
    \multirow{4}{*}{CLIP-LIT-Latent} & ($\uparrow$) PSNR & \underline{21.18} & 20.34 & 20.52 & \textbf{21.46} & 19.41    \\
                               & ($\uparrow$) SSIM & \underline{0.876} & 0.873 & 0.874 & \textbf{0.877} & 0.869 \\
                               & ($\downarrow$)  LPIPS & \underline{0.160} & 0.167 & 0.163 & \textbf{0.157} & 0.175 \\
                               & ($\downarrow$) FID & 48.17 & 47.18 & \textbf{45.28} & \underline{46.21} & 47.12 \\
    \hline
    \multirow{4}{*}{RAVE} & ($\uparrow$) PSNR & 18.46 & 18.40 & 18.21 \\
                               & ($\uparrow$) SSIM & 0.803 & 0.797 & 0.795 \\
                               & ($\downarrow$) LPIPS & 0.225 & 0.233 & 0.236 \\
                               & ($\downarrow$) FID & 57.62 & 63.00 & 64.41 \\

\bottomrule
  \end{tabular}
\end{table*}

\clearpage
\section{Training time comparison}

In Tab. \ref{tab:time} we provide quantitative comparison of training time required by different approaches to achieve decent performance on the test set.

\begin{table}[hbt!]
  \caption{Quantiative comparison of training time required by CLIP-LIT, CLIP-LIT-Latent and RAVE to achieve good performance on the test set.}
  \label{tab:time}
  \centering
  \begin{tabular}{@{}l|l|l|l@{}}
    \toprule
    Method & CLIP-LIT & CLIP-LIT-Latent & RAVE \\
    \hline
    Minutes per epoch & 95 & 94 & 42 \\ 
    Epochs to qualitative results & 6 & 6 & 2 \\
    Minutes to qualitative results  & 570 & 564 & 84 \\
  \bottomrule
  \end{tabular}
\end{table}

\section{Results on Low-Light Image Enhancement Task}

\begin{figure}[hbt!]
  \centering
      \includegraphics[width=\textwidth]{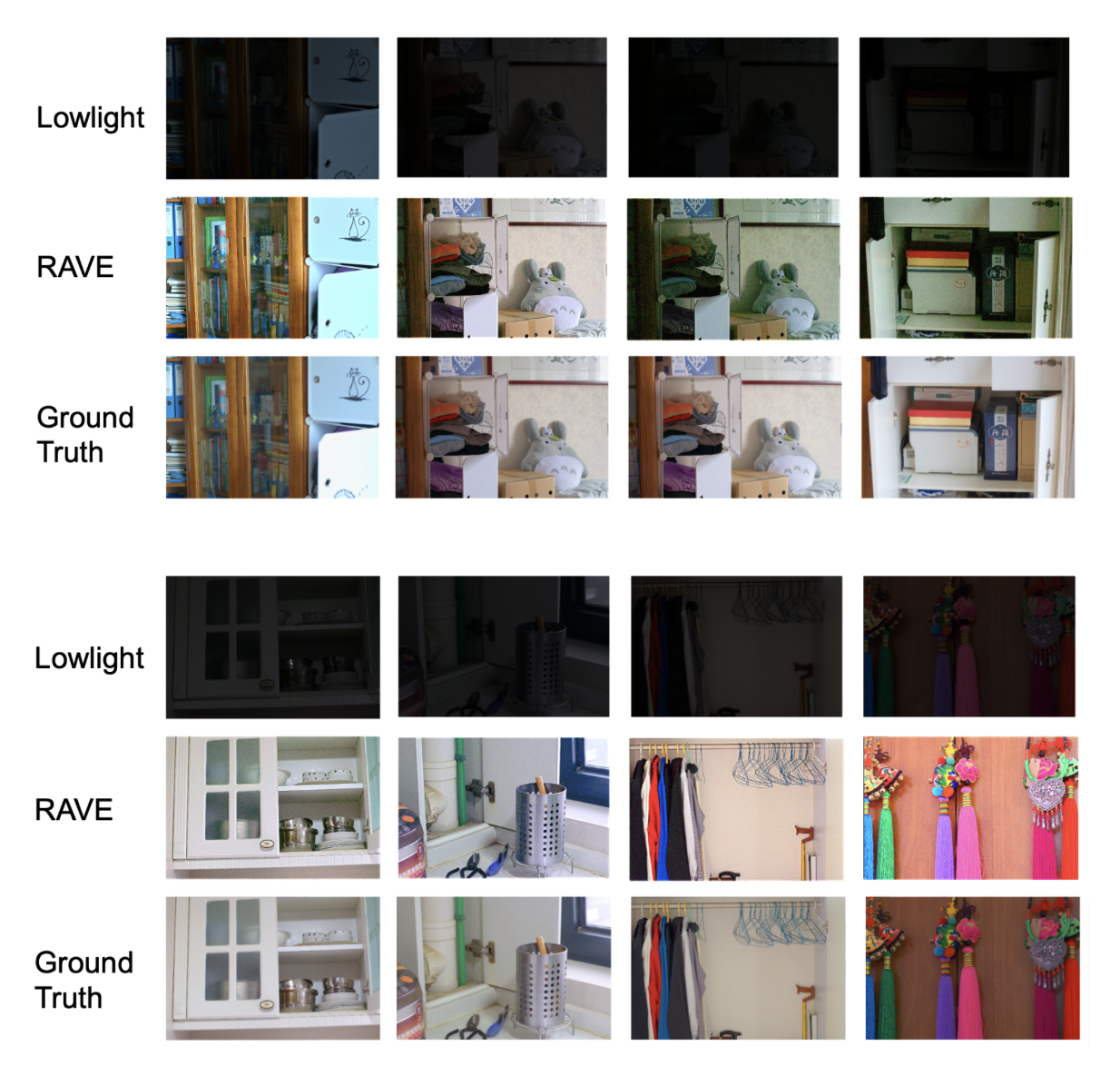}
  \caption{Visualization of RAVE performance on samples from LOL-v1 eval set. }
  \label{fig:visual-lowlight}
\end{figure}

We trained our RAVE model on a low-light image enhancement task using LOL-v1 dataset\cite{wei2018deep}. Results compared to other models are presented in Tab. \ref{tab:lol-results}. We see that RAVE does not show competitive performance to modern approaches designed specifically for the task of low-light image enhancement, though visual results obtained by RAVE are of decent quality (see Fig. \ref{fig:visual-lowlight}).

We also find vocabulary tokens, most and least similar to the $\mathbf{v}_{\text{residual}}$ calculated using LOL-v1 train set. Results are presented in Tab. \ref{tab:emb_sims_lol}. We see that for the LOL-v1 dataset vector $\mathbf{v}_{\text{residual}}$ also reflects the idea of shifting from low-light to normal light images in CLIP embedding space.

\begin{table}[hbt!]
  \caption{Best results of RAVE on LOL-v1 eval dataset compared to other lowlight image enhancement methods.}
  \label{tab:lol-results}
  \centering
  \begin{tabular}{@{}L{2.5cm}|C{2cm}C{2cm}C{2cm}@{}}
    \toprule
    Method & PSNR & SSIM & FID \\
    \hline
    RetinexNet\cite{wei2018deep} & 17.56 & 0.698 & 150.50 \\
    URetinex\cite{uretinex} & 21.33 & 0.835 & 85.59 \\
    Restormer\cite{restormer} & 22.43 & 0.823 & 78.75 \\
    Retinexformer\cite{retinexformer} & 25.16 & 0.845 & 72.38 \\
    DiffIR\cite{diffir} & 23.15 & 0.828 & 70.13 \\
    Diff-Retinex\cite{diff-retinex} & 21.98 & 0.852 & 51.33 \\
    RAVE (ours) & 17.47 & 0.509 & 132.81 \\
    
  \bottomrule
  \end{tabular}
\end{table}

\begin{table}[hbt!]
  \caption{Vocabulary tokens and their cosine similarities, which have the lowest and highest cosine similarity to the $\mathbf{v}_{\text{residual}}$ vectors calculated for LOL-v1 dataset.}
  \label{tab:emb_sims_lol}
  \centering
  \begin{tabular}{@{}ll|ll@{}}
    \toprule
    \multicolumn{2}{c|}{Lowest similarity} & \multicolumn{2}{c|}{Highest similarity}  \\
    \hline
    darkness & -0.140 & social & 0.090 \\ 
    dark & -0.126 & yotpo & 0.070 \\
    nighttime & -0.120 & homeitems & 0.070 \\
    webcamtoy & -0.117 & facilities & 0.070 \\
    candlelight & -0.114 & amarketing & 0.065 \\
    afterdark & -0.112 & njcaa & 0.063 \\
    moonlight & -0.111 & traveltips & 0.061 \\
  \bottomrule
  \end{tabular}
\end{table}
\clearpage
\newpage
\newpage

\section{Analysis of RAVE shifted}
Here we present analysis of effect of shifting the residual vector of RAVE using $\mathbf{v}_{\text{residual}}$ in the case of unpaired data.

\begin{figure}[hbt!]
  \centering
      \includegraphics[width=0.9\textwidth]{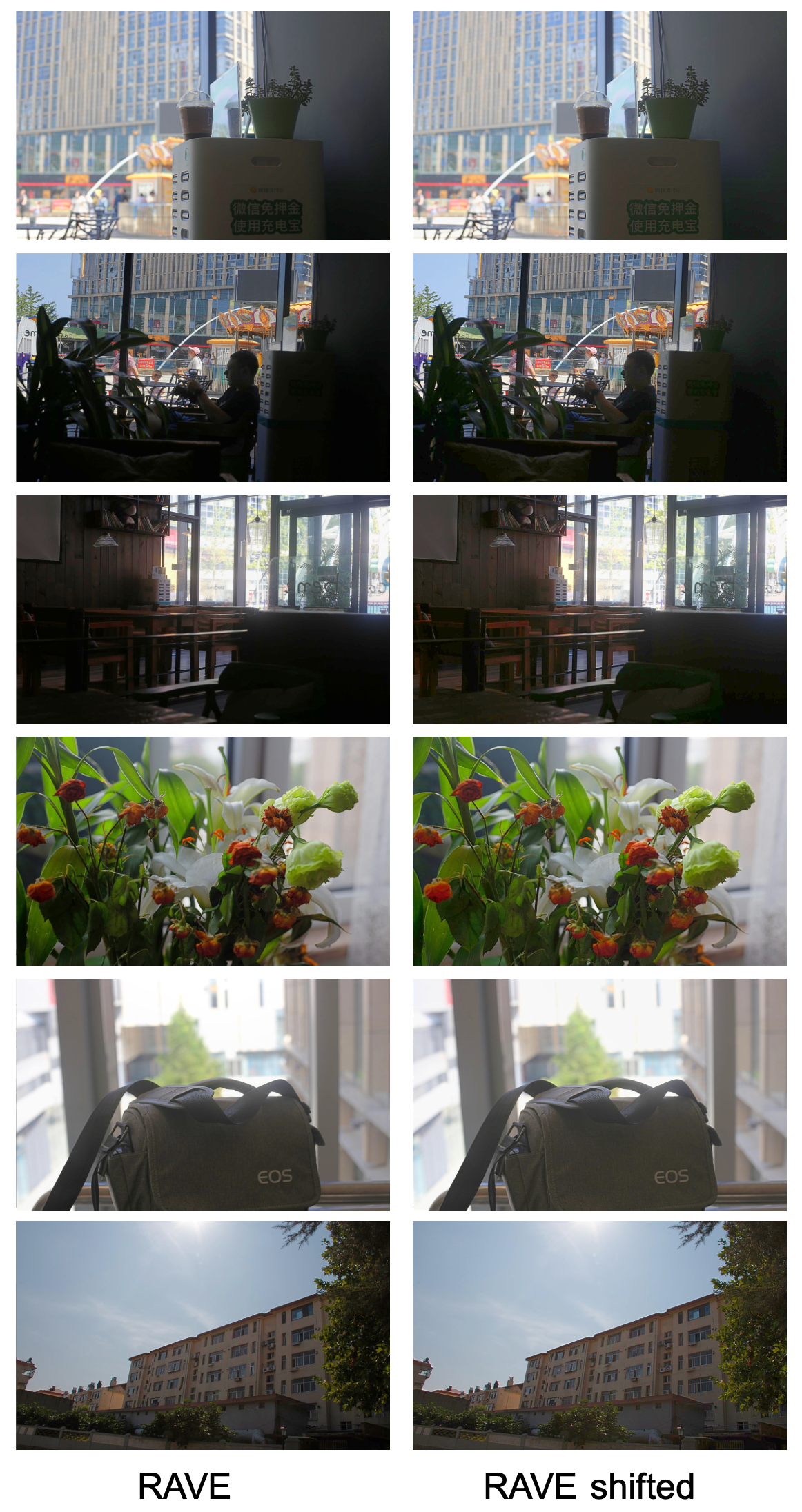}
  \caption{Visual comparison of results obtained by RAVE and RAVE (shifted version) on samples from the BAID test set. We see that RAVE shifted produces more well-lit results, while keeping advantages of RAVE, i.e. not generating over-exposed regions or artifacts.
  }
  \label{fig:rave-vs-shifted}
\end{figure}

In Tab. \ref{tab:emb_sims_shifted} we present vocabulary tokens which have most and least similar embeddings to the updated  $\mathbf{v}_{\text{residual}}$. We see that there appear tokens with meaning of "darkness" ("sunsets", "silhouette") among those which have least similar embeddings to the $\mathbf{v}_{\text{residual}}$. We also see that tokens with the most similar embeddings to $\mathbf{v}_{\text{residual}}$ now do not share any meaningful semantics. Furthermore, in Tab. \ref{tab:sim_silhouette} we show the cosine similarity between the embedding of the token "silhouette" and $\mathbf{v}_{\text{residual}}$ in case of original RAVE and RAVE shifted. We see that this similarity for the shifted $\mathbf{v}_{\text{residual}}$ is much more expressed. Finally, Fig. \ref{fig:rave-vs-shifted} presents visual comparison of images obtained using RAVE and RAVE shifted. We see that RAVE shifted produces more well-lit results, while keeping advantages of RAVE, i.e. not generating over-exposed regions or artifacts. All this supports our claim that shifting $\mathbf{v}_{\text{residual}}$ using $\mathbf{v}_{\text{add\_residual}}$ results in better guidance for the RAVE model.

\begin{table}[hbt!]
  \caption{Vocabulary tokens and their cosine similarities, which have the lowest and highest cosine similarity to the $\mathbf{v}_{\text{residual}}$ vector calculated using unpaired data and shifted using $\mathbf{v}_{\text{residual}}$.}
  \label{tab:emb_sims_shifted}
  \centering
  \begin{tabular}{@{}ll|ll@{}}
    \toprule
    \multicolumn{2}{c|}{Lowest similarity} & \multicolumn{2}{c|}{Highest similarity}  \\
    \hline
    calder & -0.138 & taeyeon & 0.093 \\ 
    sunsets & -0.134 & yoona & 0.089 \\
    apartment & -0.130 & jiu & 0.088 \\
    library & -0.128 & soyu & 0.083 \\
    basketball & -0.126 & bora & 0.073 \\
    netball & -0.126 & iu & 0.065 \\
    silhouette & -0.125 & hyo & 0.065 \\
  \bottomrule
  \end{tabular}
\end{table}

\begin{table}[hbt!]
  \caption{Vocabulary tokens and their cosine similarities, which have the lowest and highest cosine similarity to the $\mathbf{v}_{\text{residual}}$ vector calculated using unpaired data and shifted using $\mathbf{v}_{\text{residual}}$.}
  \label{tab:sim_silhouette}
  \centering
  \begin{tabular}{l|l}
    \toprule
    
    Model & Similarity \\
    \hline
     RAVE & -0.063 \\ 
     \hline
     RAVE shifted & -0.125 \\ 
  \bottomrule
  \end{tabular}
\end{table}

\clearpage
\newpage
\newpage

\section{More Visual Comparisons}

\begin{figure}[hbt!]
  \centering
      \includegraphics[width=\textwidth]{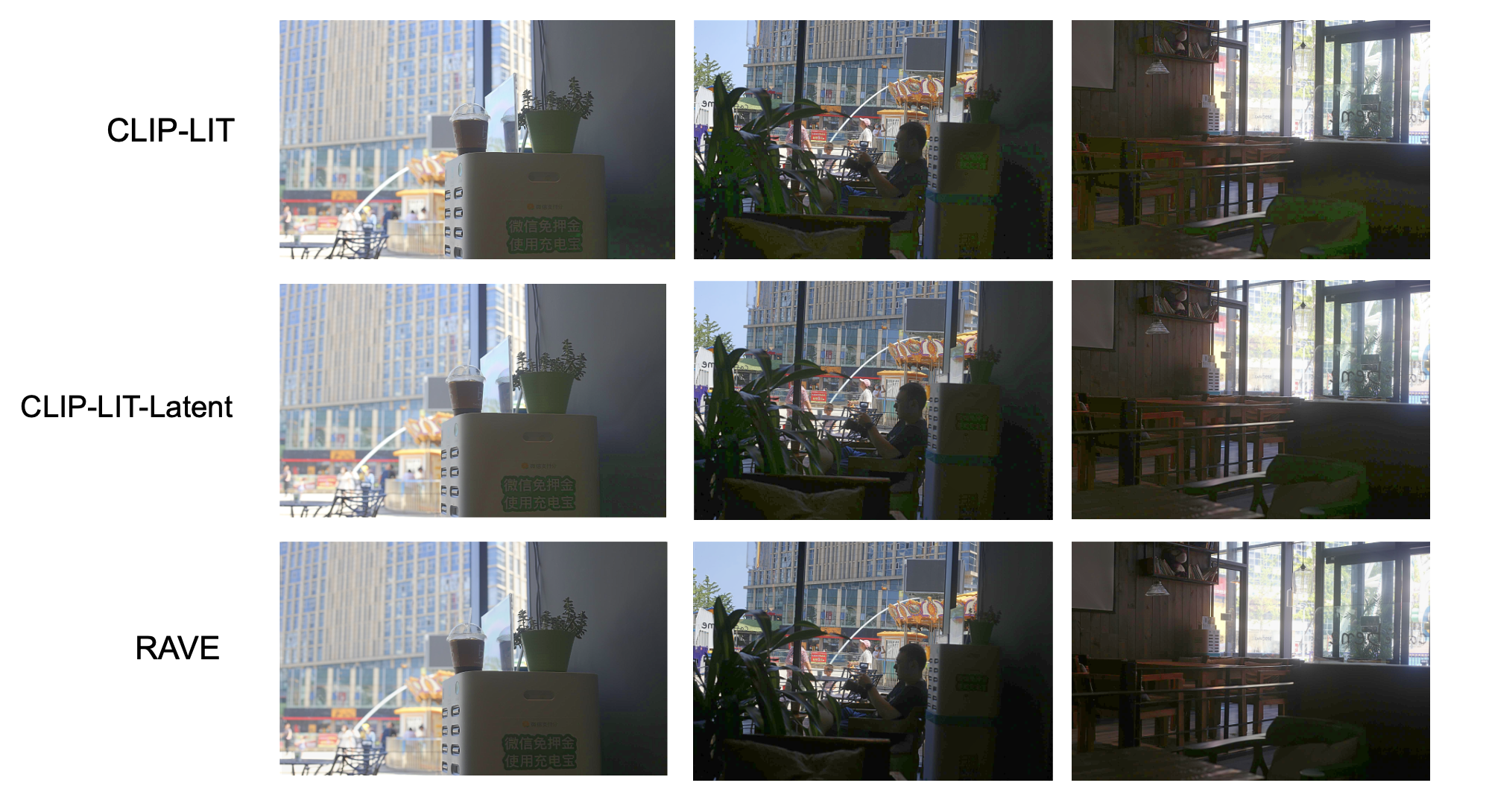}
  \caption{Visualization of performances of CLIP-LIT, CLIP-LIT-Latent and RAVE trained using \textit{unpaired} data on highly under-exposed samples from the BAID test set. We see that RAVE does not produce green artifacts in highly under-exposed areas. 
  }
  \label{fig:visual-greened}
\end{figure}

\begin{figure}[hbt!]
  \centering
      \includegraphics[width=\textwidth]{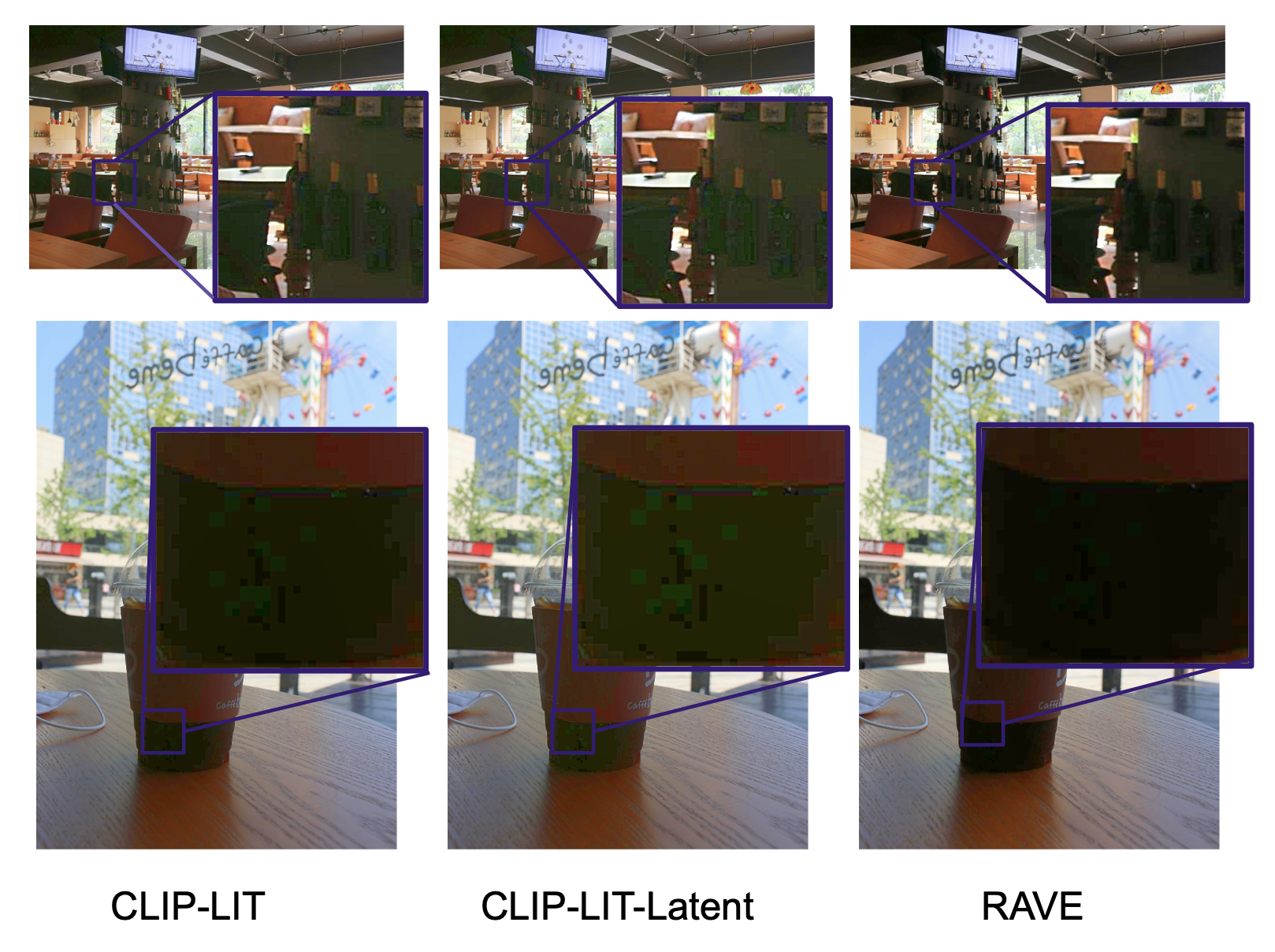}
  \caption{Visualization of performances of CLIP-LIT, CLIP-LIT-Latent and RAVE trained using \textit{unpaired} data on samples from the BAID test set. We see that RAVE does not produce green artifacts in highly under-exposed areas. 
  }
  \label{fig:visual-artifacts}
\end{figure}

\begin{figure}[hbt!]
  \centering
      \includegraphics[width=\textwidth]{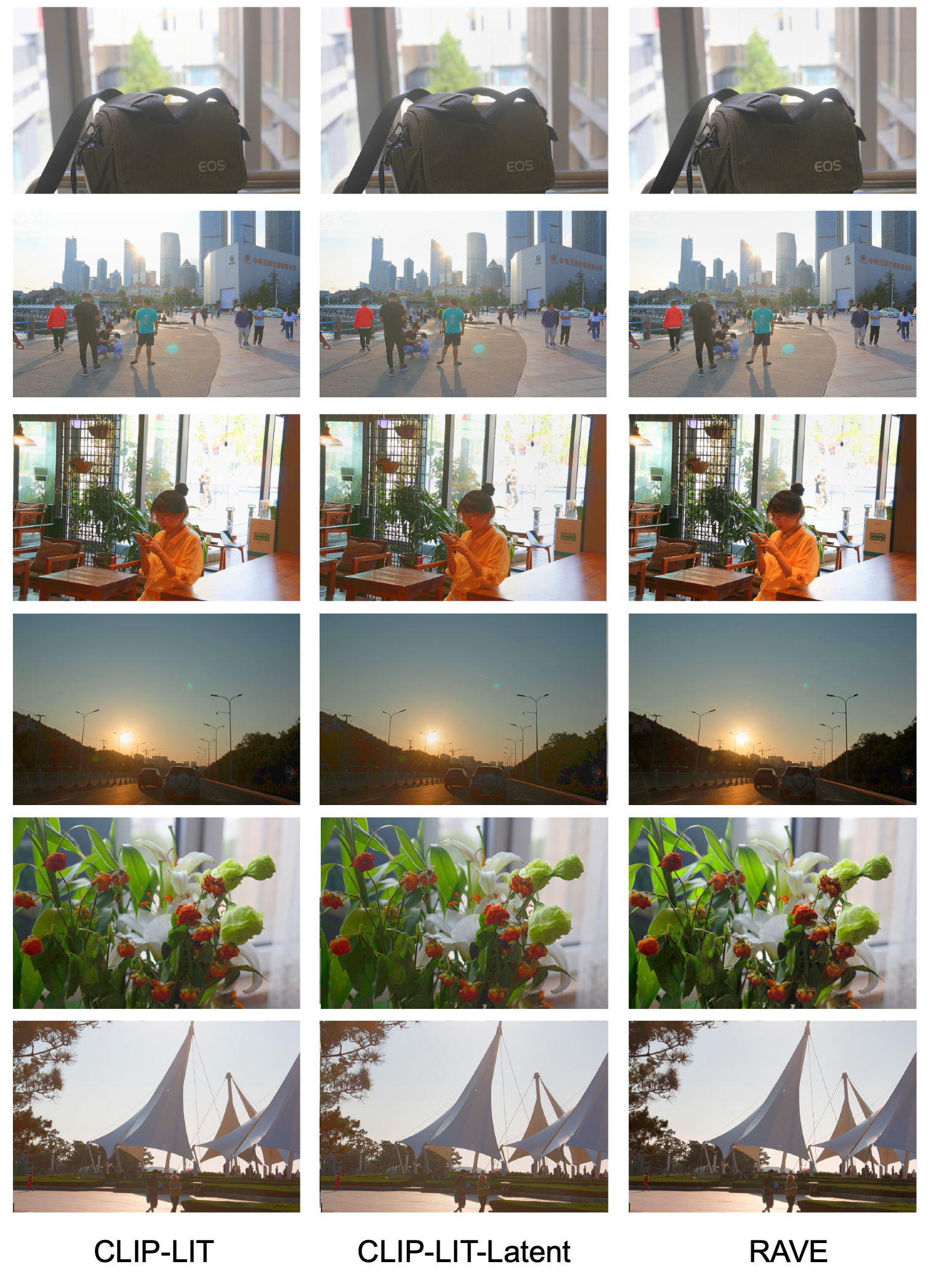}
  \caption{Visualization of results obtained by CLIP-LIT, CLIP-LIT-Latent and RAVE trained on \textit{paired} data on samples from the BAID test set. We see that RAVE in general produces images with better contrast.
  }
  \label{fig:visual-sup-1}
\end{figure}

\begin{figure}[hbt!]
  \centering
      \includegraphics[width=\textwidth]{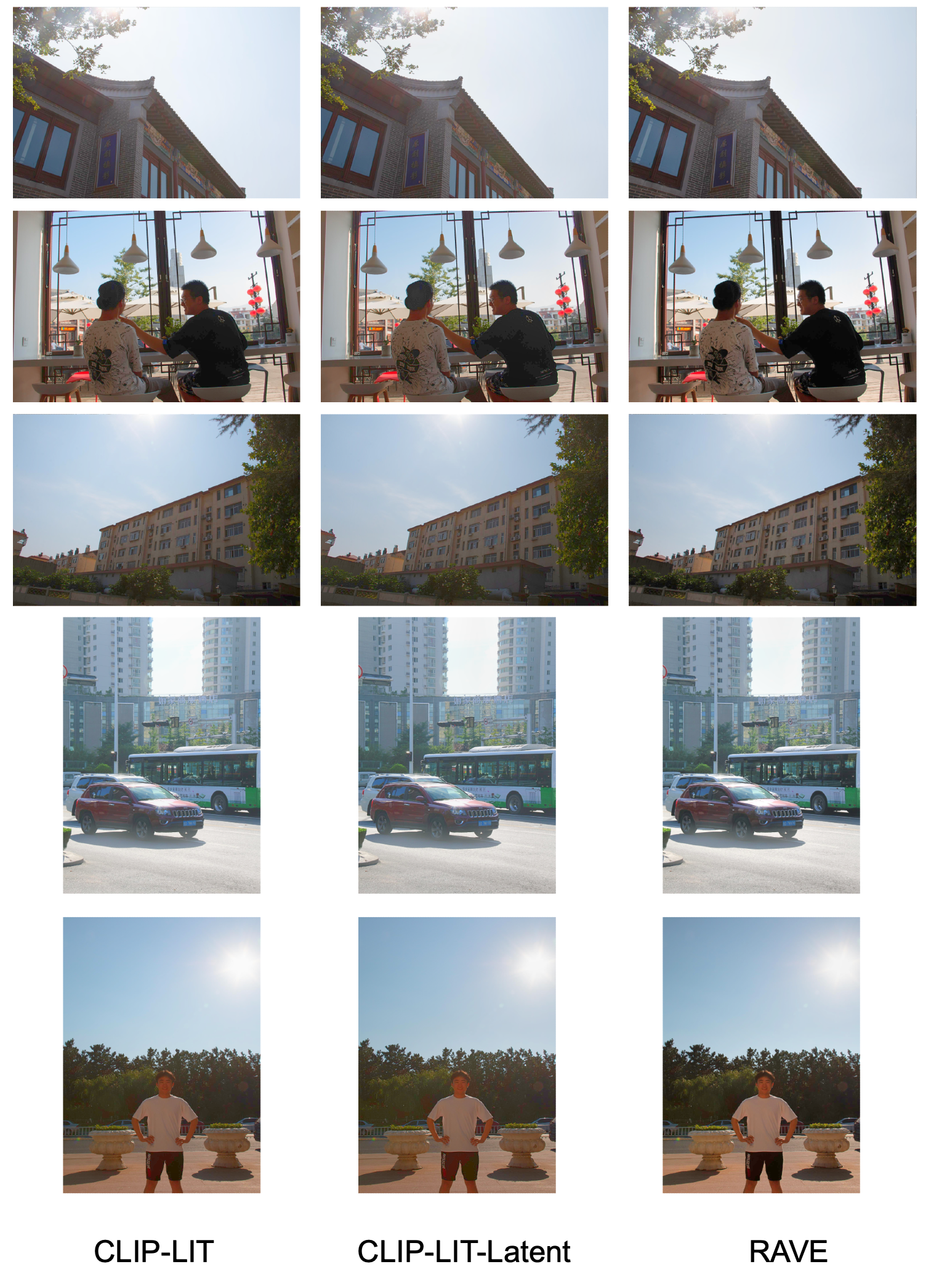}
  \caption{Visualization of results obtained by CLIP-LIT, CLIP-LIT-Latent and RAVE trained on \textit{paired} data on samples from the BAID test set. We see that RAVE in general produces images with better contrast.
  }
  \label{fig:visual-sup-2}
\end{figure}

\begin{figure}[hbt!]
  \centering
      \includegraphics[width=\textwidth]{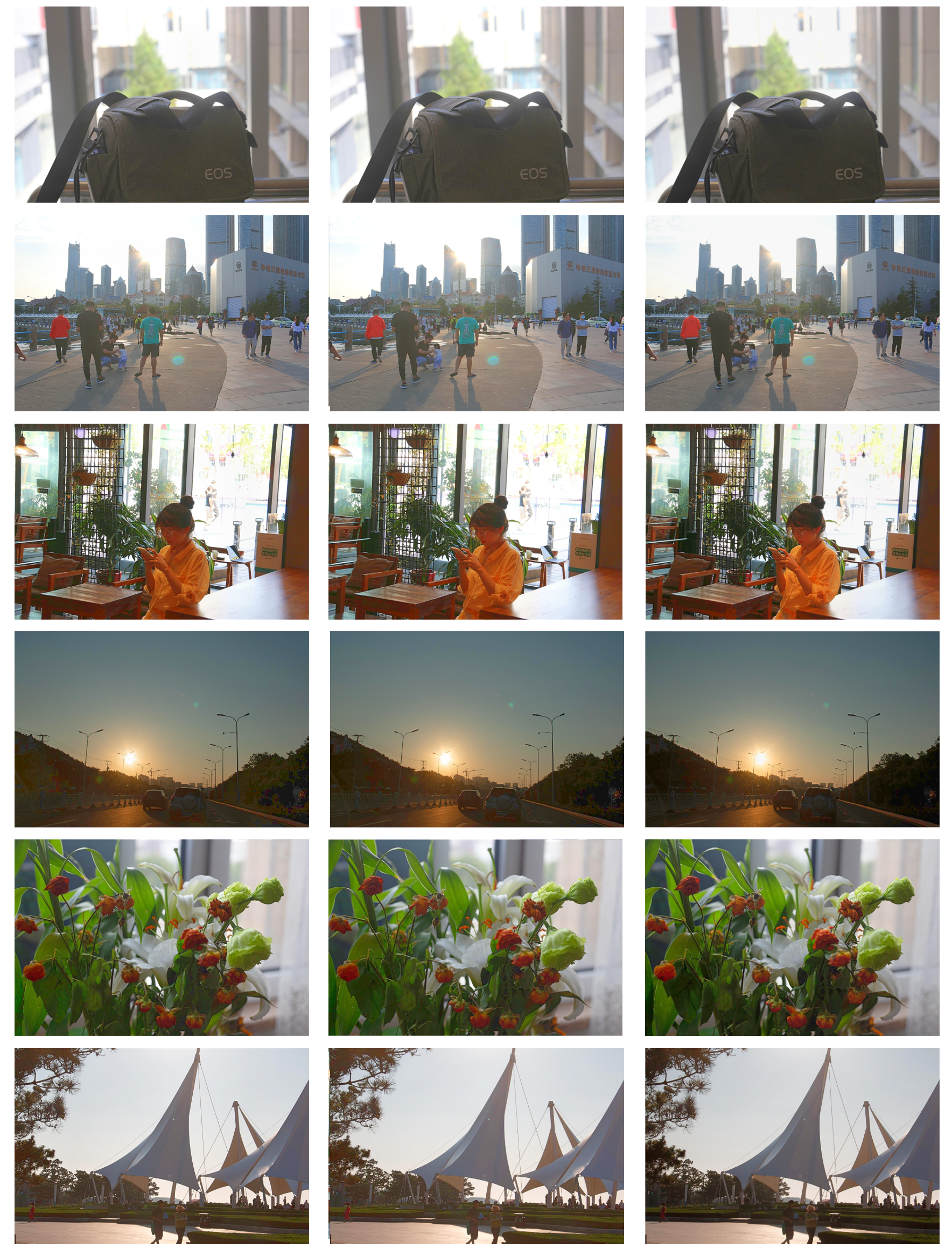}
  \caption{Visualization of results obtained by CLIP-LIT, CLIP-LIT-Latent and RAVE (shifted version) trained on \textit{unpaired} data on samples from the BAID test set.
  }
  \label{fig:visual-unsup-1}
\end{figure}

\begin{figure}[hbt!]
  \centering
      \includegraphics[width=\textwidth]{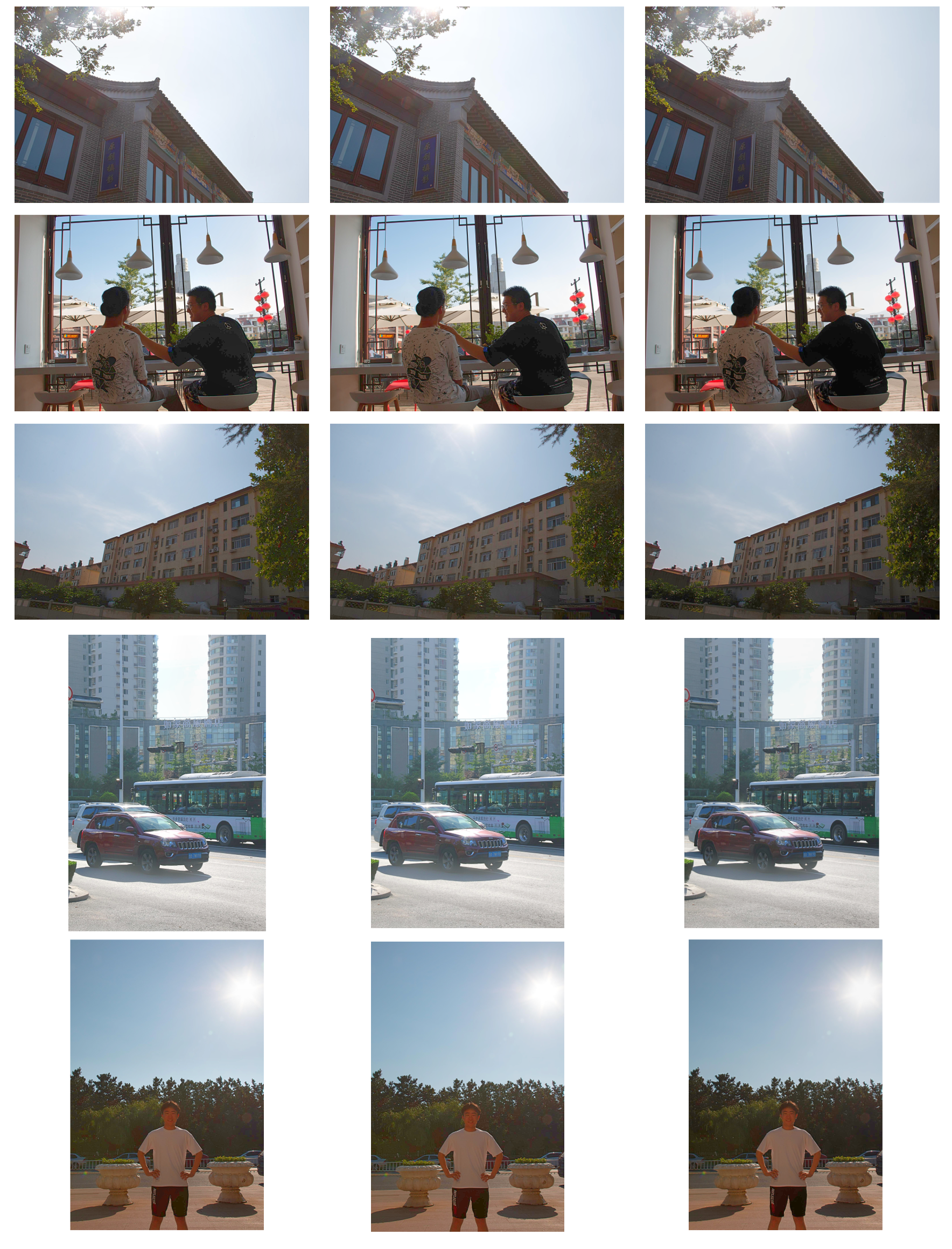}
  \caption{Visualization of results obtained by CLIP-LIT, CLIP-LIT-Latent and RAVE (shifted version) trained on \textit{unpaired} data on samples from the BAID test set.
  }
  \label{fig:visual-unsup-2}
\end{figure}

Here we provide more results on visual comparisons between CLIP-LIT, CLIP-LIT-Latent and RAVE. Figures \ref{fig:visual-sup-1} and \ref{fig:visual-sup-2} show enhanced samples from the BAID test set, obtained by models trained using \textit{paired} data. Figures \ref{fig:visual-unsup-1} and \ref{fig:visual-unsup-2} show enhanced samples from the BAID test set, obtained by models trained using \textit{unpaired} data. Further, in Figures \ref{fig:visual-greened} and \ref{fig:visual-artifacts} we show enhanced results on samples from the BAID dataset which have highly under-exposed areas, obtained by model trained using \textit{unpaired} data. We see that RAVE, unlike CLIP-LIT or CLIP-LIT-Latent, avoids getting visual artifacts on those areas.

\clearpage
\newpage
\bibliographystyle{splncs04}
\bibliography{main}